\documentclass{article} % For LaTeX2e
\usepackage[dvipsnames]{xcolor}     % 支持自定义颜色（如 gray!10, pink!20, red, gre

\usepackage{iclr2026_conference,times}

\usepackage{amsmath,amsfonts,bm}

\def\eqref#1{equation~\ref{#1}}
\def\1{\bm{1}}

\DeclareMathAlphabet{\mathsfit}{\encodingdefault}{\sfdefault}{m}{sl}
\SetMathAlphabet{\mathsfit}{bold}{\encodingdefault}{\sfdefault}{bx}{n}

\usepackage{hyperref}
\usepackage{url}
\usepackage{multirow}   % 支持 \multirow
\usepackage{booktabs}   % 支持 \toprule, \midrule, \bottomrule, \cmidrule
\usepackage{colortbl}   % 支持 \rowcolor
\usepackage{graphicx}   % 支持 \resizebox
\usepackage{array}      % 加强 tabular 环境，可自定义列格式
\usepackage[labelfont=bf]{caption}   
\usepackage{enumitem}
\usepackage{wrapfig}    % 支持 wrapfigure 环绕浮动\
\usepackage{pifont}
\usepackage{subcaption}
\usepackage[most]{tcolorbox}
\usepackage{xspace}

\definecolor{brandblue}{rgb}{0.54, 0.7, 1}

\newcommand{\tooltoken}[1]{\texttt{#1}}
\newcommand{\method}{AdaReasoner}
\usepackage{enumitem}

\newcommand{\hfemoji}{%
  \raisebox{-0.20\height}{%
    \hspace{0.1em}
    \includegraphics[height=1em]{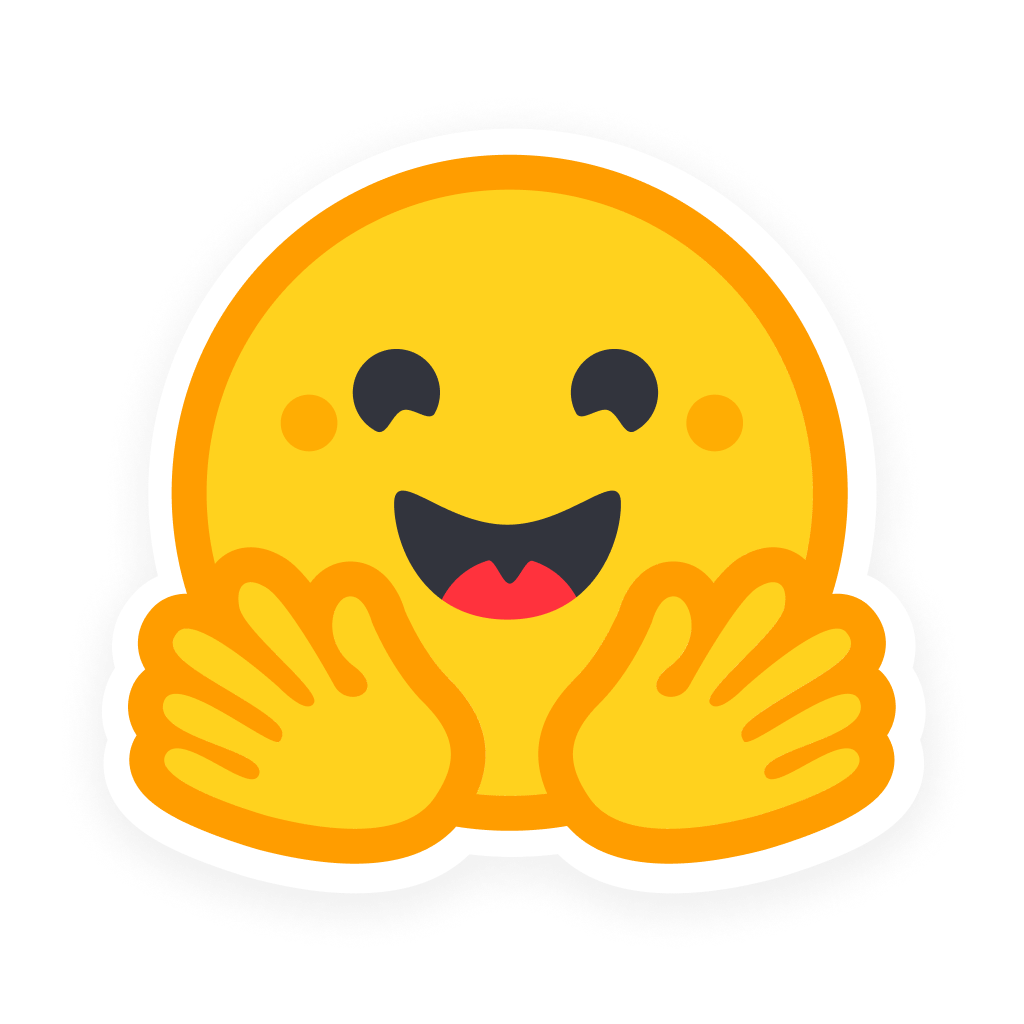}%
  }\xspace
}

\newcommand{\githubemoji}{%
  \raisebox{-0.20\height}{%
    \hspace{0.15em}
    \includegraphics[height=0.9em]{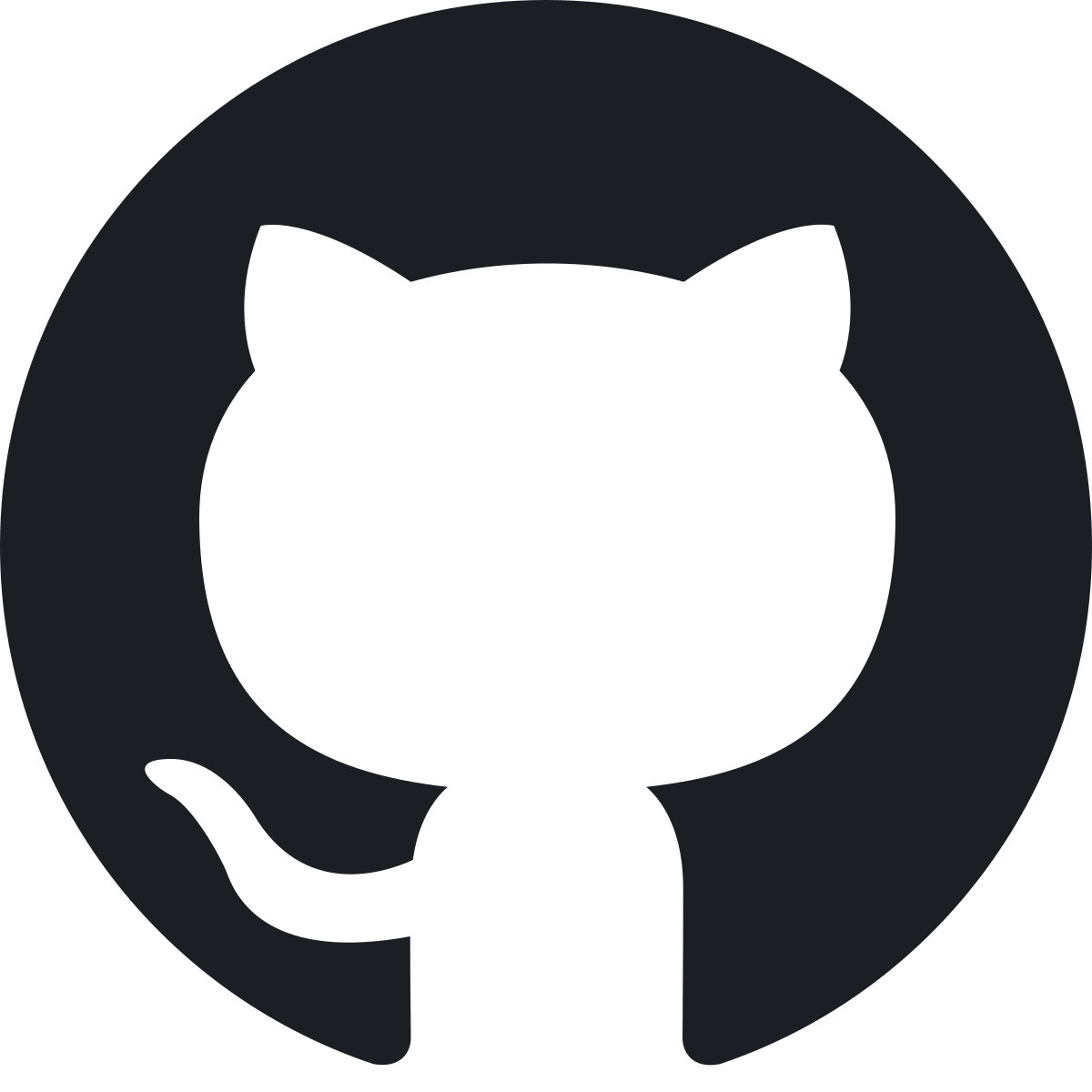}%
  }\xspace
}

\newcommand{\houseemoji}{%
  \raisebox{-0.20\height}{%
    \hspace{0.15em}
    \includegraphics[height=0.9em]{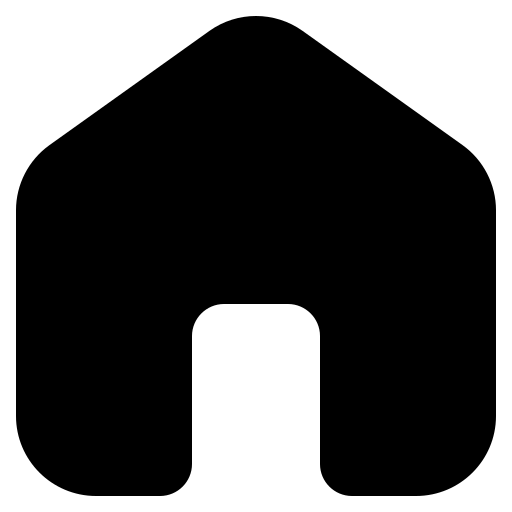}%
  }\xspace
}

\definecolor{promptframecolor}{RGB}{102,102,178}
\definecolor{promptbackcolor}{RGB}{244,251,254}

\newtcolorbox{promptbox}[2][]{
  colback=promptbackcolor, 
  colframe=promptframecolor, 
  fonttitle=\bfseries, 
  coltitle=white, 
  title=#2,
  #1
}
\usepackage{xcolor}
\usepackage{colortbl}
 \usepackage{calc}
\usepackage{booktabs}
\definecolor{customblue}{rgb}{0.2, 0.3, 0.8}
\definecolor{customgreen}{rgb}{0.1, 0.6, 0.3}
\usepackage{pifont}
\usepackage{fontawesome5}

\definecolor{lightblue}{RGB}{173, 216, 230}
\definecolor{lightgray}{gray}{0.9}
\definecolor{lightgreen}{RGB}{144, 238, 144}
\definecolor{lightred}{RGB}{255, 182, 193}
\usepackage{wrapfig}

\usepackage{tikz}

\definecolor{skyblue}{RGB}{0, 102, 204}

\usepackage{xcolor} 
\usepackage{graphicx} 
\usepackage{fontawesome5}
\usepackage{listings}
\usepackage{multirow}
\usepackage{tabularx}
\usepackage{booktabs}
\usepackage{pifont}
\usepackage{circledsteps}
\usepackage{wrapfig}

\usepackage{float}
\tcbset{
  highlightbox/.style={
    enhanced,
    colback=gray!10,
    colframe=gray!50,
    boxrule=0.5pt,
    arc=2pt,
    left=6pt, right=6pt, top=6pt, bottom=6pt
  }
}

\definecolor{wkred}{RGB}{255, 190, 190}
\definecolor{wkblue}{RGB}{210, 230, 250}
\definecolor{wkgreen}{RGB}{226,240,217}

\definecolor{mygray}{gray}{0.4}
\definecolor{lightgray}{rgb}{0.9, 0.9, 0.9}

\usepackage{etoolbox, subcaption}
\usepackage[font=footnotesize,labelfont=bf,aboveskip=1pt,belowskip=-10pt]{caption}

\newdimen\abovecrulesep
\newdimen\belowcrulesep
\makeatletter
\patchcmd{\@@@cmidrule}{\aboverulesep}{\abovecrulesep}{}{}
\patchcmd{\@xcmidrule}{\belowrulesep}{\belowcrulesep}{}{}

\definecolor{demphcolor}{RGB}{144, 144, 144}
\definecolor{mygray}{gray}{0.4}
\definecolor{lightgray}{rgb}{0.9, 0.9, 0.9}

\newlength\savewidth

\newcommand{\tablestyle}[2]{\setlength{\tabcolsep}{#1}\renewcommand{\arraystretch}{#2}\centering\footnotesize}
\makeatletter\renewcommand\paragraph{\@startsection{paragraph}{4}{\z@}{.5em\@plus1ex\@minus.2ex}{-.5em}{\normalfont\normalsize\bfseries}}
\makeatother

\newcolumntype{C}[1]{>{\centering\arraybackslash}p{#1}}
\newcolumntype{R}[1]{>{\raggedleft\arraybackslash}p{#1}}
\newcolumntype{L}[1]{>{\raggedright\arraybackslash}p{#1}}

\preto\align{\small}
\expandafter\preto\csname align*\endcsname{\small}
\preto\equation{\par\nobreak\small\noindent}

\usepackage{graphicx}
\usepackage{adjustbox}

\usepackage[utf8]{inputenc}
\usepackage{booktabs}
\usepackage{multirow}
\usepackage{siunitx}
\usepackage{colortbl}
\usepackage{xcolor}
\usepackage{enumitem}

\usepackage{ulem}
\renewcommand{\emph}[1]{\uline{#1}}

\title{
    \adjustbox{raise=-0.15\height}{
        \includegraphics[width=2.0em]{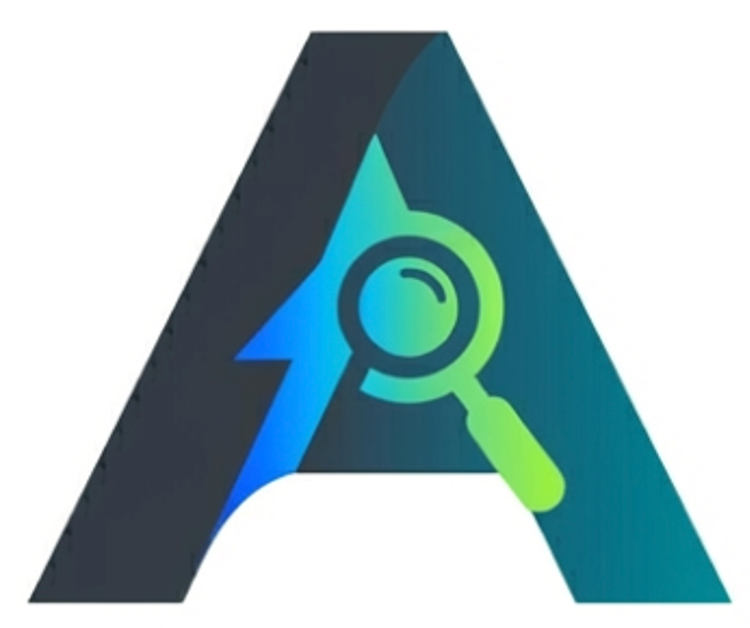}
    }
    \hspace{-0.68em}daReasoner: Dynamic Tool  Orchestration for Iterative Visual Reasoning
}

\author{
\hspace{-0.25em}\textbf{Mingyang Song}\thanks{Equal contribution. \quad $^{\dagger}$Equal Advisory Contribution}~~$^{,1}$,
\textbf{Haoyu Sun}$^{*,2}$,
\textbf{Jiawei Gu}$^{*,3}$,
\textbf{Linjie Li}$^{*,4}$,
\textbf{Luxin Xu}$^{5}$, \\
\textbf{Ranjay Krishna}$^{4,\dagger}$, 
\textbf{Yu Cheng}$^{6,\dagger}$ \\
\\
$^{1}$Fudan University,\;
$^{2}$Tongji University,\;
$^{3}$National University of Singapore, \\
$^{4}$University of Washington,\;
$^{5}$University of Electronic Science and Technology of China,\\
$^{6}$The Chinese University of Hong Kong
\\[2mm]
\houseemoji~{Homepage:}~\href{https://adareasoner.github.io/}{\texttt{https://adareasoner.github.io}}\\
\githubemoji~{Code: }~\href{https://github.com/ssmisya/AdaReasoner}{\texttt{https://github.com/ssmisya/AdaReasoner}}\\
\hfemoji~{Models and Data:}~\href{https://huggingface.co/AdaReasoner}{\texttt{https://huggingface.co/AdaReasoner}}\\
}

\iclrfinalcopy
\begin{document}

\maketitle

\begin{abstract}

When humans face problems beyond their immediate capabilities, they rely on tools, providing a promising paradigm for improving visual reasoning in multimodal large language models (MLLMs).
Effective reasoning, therefore, hinges on knowing which tools to use, when to invoke them, and how to compose them over multiple steps, even when faced with new tools or new tasks. We introduce \textbf{AdaReasoner}, a family of multimodal models that learn tool use as a general reasoning skill rather than as tool-specific or explicitly supervised behavior. AdaReasoner is enabled by (i) a scalable data curation pipeline exposing models to long-horizon, multi-step tool interactions; (ii) Tool-GRPO, a reinforcement learning algorithm that optimizes tool selection and sequencing based on end-task success; and (iii) an adaptive learning mechanism that dynamically regulates tool usage. Together, these components allow models to infer tool utility from task context and intermediate outcomes, enabling coordination of multiple tools and generalization to unseen tools. Empirically, AdaReasoner exhibits strong tool-adaptive and generalization behaviors: it autonomously adopts beneficial tools, suppresses irrelevant ones, and adjusts tool usage frequency based on task demands, despite never being explicitly trained to do so. These capabilities translate into state-of-the-art performance across challenging benchmarks, improving the 7B base model by +24.9\% on average and surpassing strong proprietary systems such as GPT-5 on multiple tasks, including VSP and Jigsaw.
\end{abstract}

\section{Introduction}

Humans often rely on external tools to solve the complex reasoning problems that go beyond what they can handle through internal thinking alone~\citep{Extended_mind_theory}. This offers a valuable perspective for improving the visual reasoning capabilities of Multimodal Large Language Models (MLLMs). 
By offloading perceptual and intermediate computation to external tools, the model can more effectively handle \emph{fine-grained visual perception}~\citep{tong2024eyes} and \emph{multi-step reasoning}~\citep{emma}, where precise intermediate verification and long-horizon planning are crucial.
In this setting, adaptive tool usage becomes particularly important. MLLMs should learn not only how to use tools, but also when to use them and how to coordinate multiple tools over several steps, including tools it has never seen before.

However, unlike humans who can flexibly decide when to resort to tools and which tools to use, current models still struggle to master adaptive tool usage.
Early SFT- and prompt-based approaches~\citep{ma2024taco,hu2024visualsketchpad} explored multi-tool usage, but largely relied on rigid, pre-defined invocation patterns rather than autonomous and adaptive planning. More recent RL-based methods, such as DeepEyes~\citep{deepeyes} and Pixel-Reasoner~\citep{su2025pixelreasoner}, have improved perceptual reasoning through crop-based search strategies. However, they are typically constrained to single-tool trajectories or fixed interaction loops.
As a result, a critical gap remains: existing methods lack the ability to \textit{adaptively plan and coordinate} diverse tools in a flexible and task-aware manner. They do not treat the decision of \emph{what} tools to use, \emph{when} to use them, and \emph{how} to combine them as a core component of multimodal reasoning. Moreover, because these approaches are not explicitly designed for adaptability, their tool-use policies tend to be brittle, exhibiting limited generalization to unseen tools or novel task distributions beyond their training scope.

\begin{figure}[t]
  \centering
  \includegraphics[width=0.98\linewidth]{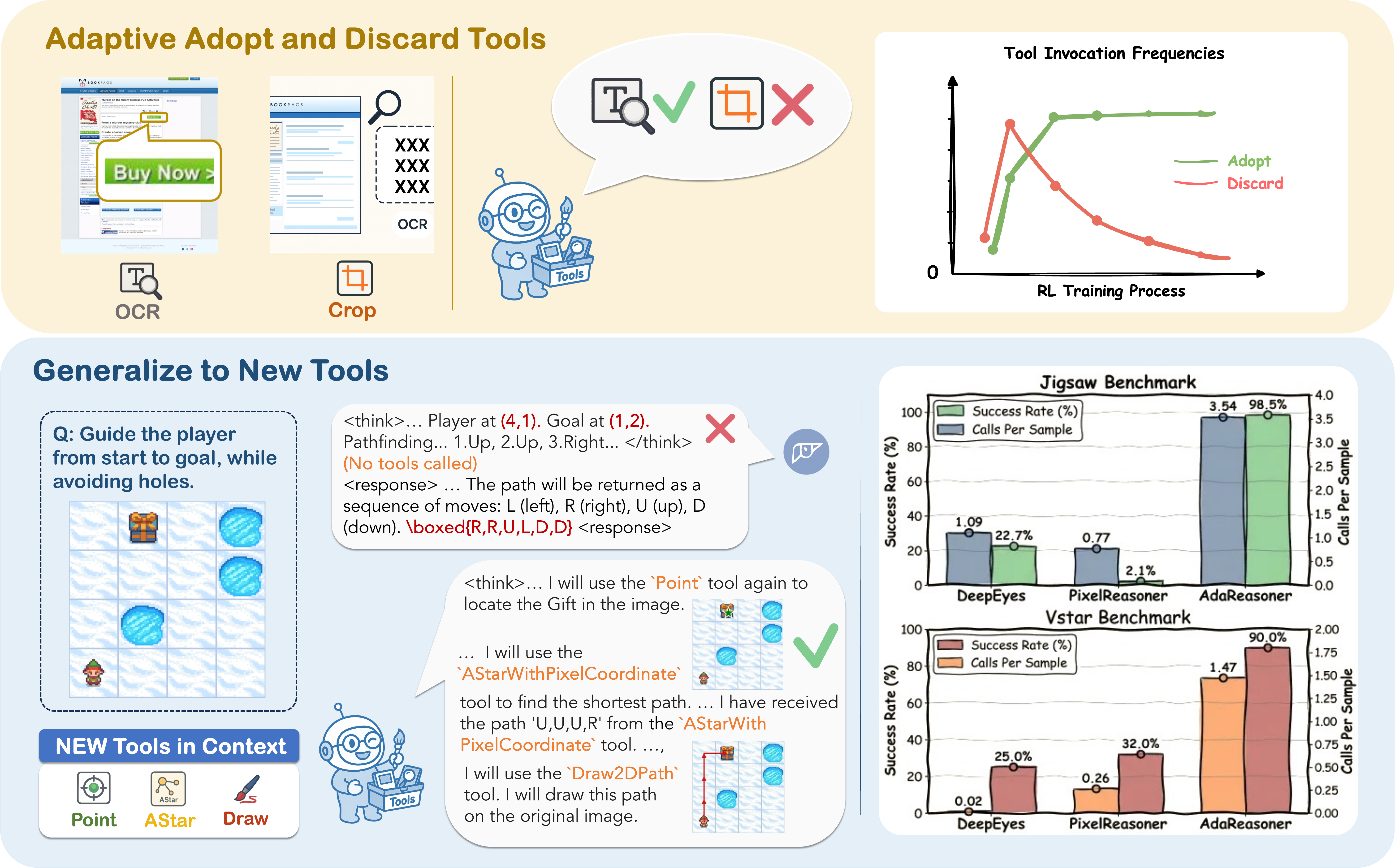}
  \caption{\textbf{\method~performs adaptive and generalized tool-using.} 
  }
  \label{fig:overview}
\end{figure}

We present \textbf{AdaReasoner}, a tool-aware reasoning model designed to overcome the limitations of rigid, single-tool paradigms and poor generalization by enabling adaptive, multi-turn tool planning.
Our framework is built upon three key innovations.
\textbf{First}, to establish a robust foundation, we introduce a new \textit{data curation pipeline} that automatically synthesizes high-quality, multi-turn trajectories. 
\textbf{Second}, we further refine this policy using a tailored \textit{Tool GRPO (TG)} paradigm optimized for long-horizon strategic planning. This training strategy enables the model to reason over extended interaction sequences and make more coherent tool-use decisions. \textbf{Finally}, integrated within both the TC and TG stages is our proposed \textit{adaptive learning method (ADL)}, which is specifically engineered to decouple tool-use logic from specific tasks, thereby significantly enhancing the model's generalizability to unseen domains. 
As illustrated in Figure~\ref{fig:overview}, this active reasoning cycle enables the agent to not only extract visual evidence but to dynamically transform it, yielding a deeper and more robust multimodal understanding.

Through adaptive tool interaction, AdaReasoner achieves substantial performance gains across diverse benchmarks. 
In particular, the 7B variant attains an average improvement of \textbf{+24.9\%}, while also surpassing strong proprietary models -- outperforming Claude Sonnet~4 and GPT-5 on multiple tasks. Beyond accuracy, as shown in Figure~\ref{fig:overview}, AdaReasoner also exhibits \textit{\textbf{self-adaptive tool-use behaviors}}. It learns to select effective tools, discard irrelevant ones, and regulate their use according to task demands and feedback, revealing strong flexibility and generalization. Moreover, AdaReasoner is able to generalize its tool-planning capability to unseen tasks and new tool definitions, maintaining both a high frequency and strong accuracy of tool usage even when encountering new tasks.
These findings offer a compelling answer to the long-standing question of which tools should be included and how models should learn to use them, suggesting that with proper training, MLLMs can autonomously curate tool-use strategies from a broad candidate set and extend their visual reasoning capacity in a goal-directed manner. In summary, our main contributions are as follows:

\vspace{-5pt}
\begin{itemize}[leftmargin=*,itemsep=5pt]

    \item We propose a comprehensive method for developing tool-augmented models, built upon three core innovations: a data curation recipe for multi-turn tool planning, an RL framework for multi-turn tool interaction, and an adaptive learning method that can enhance model's generalizability.

    \item Based on our method, we introduce \textbf{AdaReasoner}, a new family of state-of-the-art models for complex tool planning.  AdaReasoner exhibits self-adaptive behaviors, learning to autonomously \textbf{adopt} beneficial tools, \textbf{discard} irrelevant ones, and \textbf{modulate} its usage frequency, while maintaining strong generalization to unseen tool definitions and novel tasks.

    \item Our \method~achieves significant gains over their base counterparts and delivers performance that is competitive with, or superior to, leading proprietary models like GPT-5 and Claude Sonnet~4 on structured-reasoning tasks. This establishes that our methodology can elevate smaller, open-source models to the state-of-the-art.

\end{itemize}

\section{Method}

\subsection{Preliminary}

\begin{figure}[t]
    \centering
    \includegraphics[width=1\linewidth]{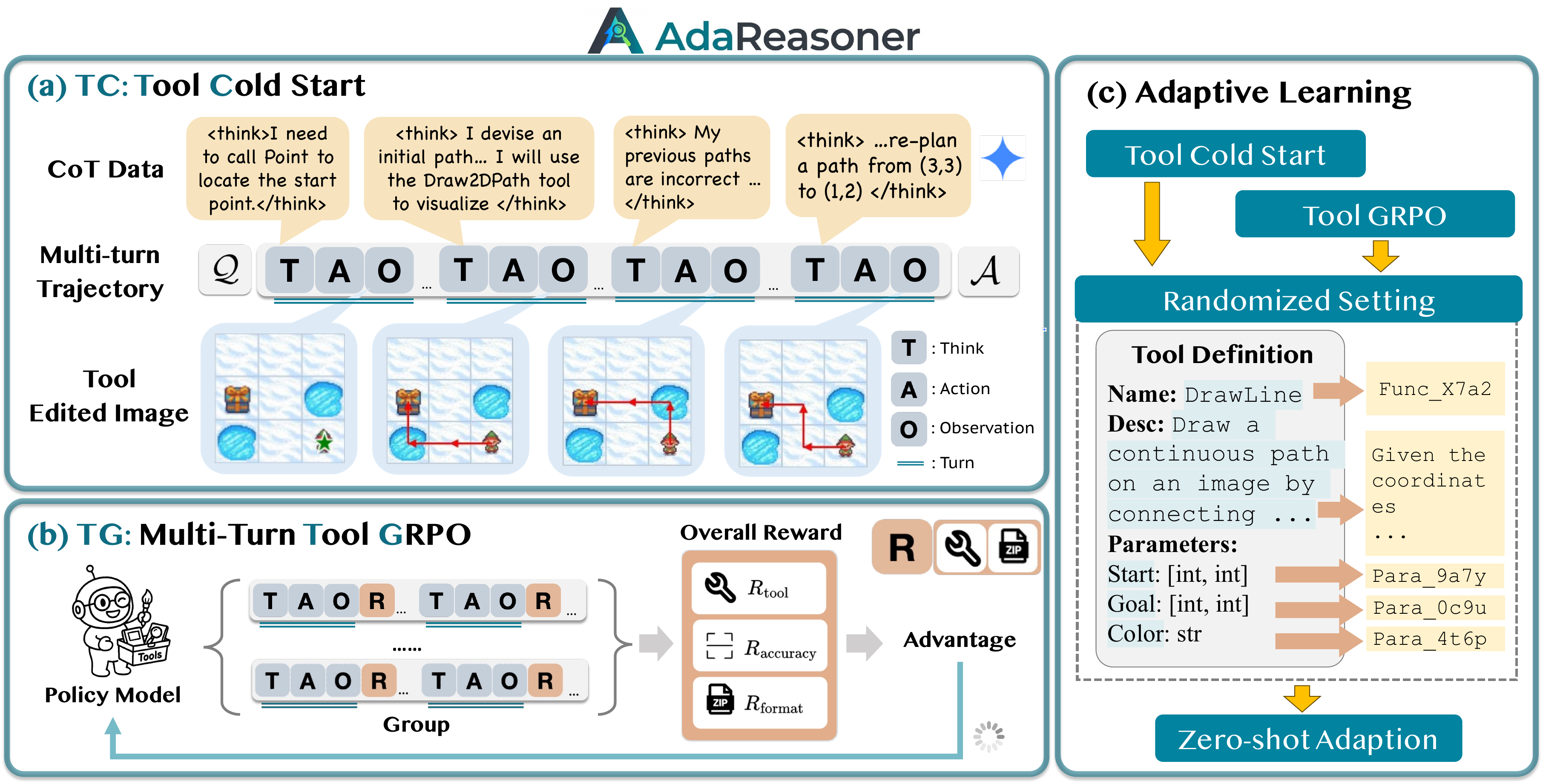}
    \caption{
    An overview of our \method~framework. The pipeline consists of two main stages: 
    (a) the \textit{Tool Cold Start (TC)} phase, where trajectories are carefully constructed to support multi-turn reasoning; and 
    (b) the \textit{Tool GRPO (TG)} phase, where the policy is further refined via reinforcement learning guided by our adaptive, multi-turn reward.
    In addition, the \textit{Adaptive Learning} method (c) can be applied throughout both the TC and TG stages, enabling improved generalization across tasks and tool configurations. }
    \label{fig:main_fig}
\end{figure}

\paragraph{Problem Formulation}
As shown in figure \ref{fig:main_fig}, we formalize tool-augmented multimodal reasoning as a sequential decision-making process. 
An MLLM represented as a policy $\pi_\theta$ parameterized by weights $\theta$, is tasked with solving a problem by generating a reasoning trajectory $\tau$. 
The policy is equipped with access to a predefined set of visual tools $T = \{t_1, \dots, t_n\}$.

A trajectory $\tau$ is a sequence of state-action-observation tuples that represent the model's step-by-step reasoning process:
\begin{equation}
    \tau = \{ (s_0, a_0, o_0), (s_1, a_1, o_1), \dots, (s_T, a_T, o_T) \}
\end{equation}
Here, $s_t$ denotes the problem state, $a_t \in \mathcal{T}$ is a tool-calling action encapsulated by special tokens, and $o_t$ is the resulting observation from the tool's execution. 
Each action $a_t$ induces a transition from state $s_t$ to $s_{t+1}$ based on the new information in $o_t$:
\begin{equation}
    s_0 \xrightarrow{a_0} s_1 \xrightarrow{a_1} s_2 \xrightarrow{a_2} \dots \xrightarrow{a_T} s_{T+1}
\end{equation}

\paragraph{Visual Tools}

Our \method~framework is built upon a diverse and powerful suite of visual tools, which it executes and integrates directly into the reasoning process. This toolset is intentionally designed to cover three core reasoning functions: \textbf{perception} (e.g., \textsc{Point}, \textsc{OCR}), \textbf{manipulation} (e.g., \textsc{DrawLine}, \textsc{InsertImage}), and \textbf{calculation} (e.g., \textsc{AStar}). Furthermore, this suite seamlessly integrates both lightweight, offline tools for immediate execution and computationally intensive, expert-model-based online tools. These foundational capabilities are summarized in Table~\ref{tab:vision-tools}, with detailed specifications for each tool provided in Appendix~\ref{apdx:tool-definition}.

\begin{table*}[h]
\centering
\tablestyle{4pt}{1.1}
\resizebox{\textwidth}{!}{%
\begin{tabular}{llll}
\toprule
\textbf{Tool}                    & \textbf{Description}              & \textbf{Arguments}               & \textbf{Tool Output}           \\
\midrule
\textsc{Point} & Point to a target object & Image + Description & Point coordinates \\
\textsc{Draw2DPath} & Draw a path using directional commands & Image + Start + Directions & Image with a line \\
\textsc{AStar} & Use A* to find the shortest obstacle-free path & Start + Goal + Obstacle & Shortest path \\
\textsc{DetectBlackArea} & Detect pure black areas in an image  & Image & Bounding boxes of black areas \\
\textsc{InsertImage} & Insert image into base at bounding box position & Image + Coordinates + Insert & Combined image \\
\textsc{OCR} & Extracts and localizes text from the image & Image & Text with their bounding box \\
\textsc{Crop} & Crop a region and augment it & Image + Coordinates & Cropped Image \\
\bottomrule
\end{tabular}}

\caption{Visual tools integrated within \method. We illustrate their arguments, outputs, and core functions description. More detailed descriptions of our tools are presented in Appendix \ref{apdx:tool-definition}.}
\label{tab:vision-tools}
\end{table*}

\subsection{High-Quality Trajectory Data Curation}

As illustrated in Figure~\ref{fig:main_fig}a, our data curation follows a unified, three-stage process designed to generate high-fidelity, human-like reasoning trajectories.

\textbf{Abstract Trajectory Design} ~~~First, for each task, we manually design an abstract, optimal problem-solving blueprint. For example, the \textbf{VSP} trajectory follows a perception-planning-verification logic, \textbf{Jigsaw} mimics an iterative trial-and-error process, and \textbf{GUIQA} involves a focus-then-extract strategy. However, to ensure the model develops true robustness beyond simply following these ``perfect" paths, we deliberately incorporate two critical types of complex scenarios:
\vspace{-5pt}
\begin{itemize}[leftmargin=*,itemsep=2pt]
    \item \textbf{Reflection and Backtracking:} We include trajectories designed to encourage a process of trial and verification. These feature explicit self-correction steps where the model must reflect on a sub-optimal outcome and backtrack, teaching it to actively validate its own hypotheses and learn from intermediate failures.
    
    \item \textbf{Explicit Tool Failure:} To prevent over-reliance on external tools, we introduce cases where tools fail or return erroneous results. In these scenarios, after recognizing that a tool is not providing a useful output, the trajectory prompts the model to fall back on its own intrinsic capabilities to generate a ``best-effort'' answer, ensuring it develops a resilient, dual-strategy approach.
\end{itemize}

\textbf{Tool Calling supplements}~~~
Subsequently, we ground these abstract blueprints by programmatically executing the tool calls to populate them with concrete, real-world inputs and outputs.

\textbf{CoT Data Generation} ~~~Finally, we leverage a powerful LLM to generate the corresponding Chain-of-Thought (CoT) reasoning that connects each step. This process yields a final dataset of rich, tool-augmented trajectories that teach the model not just \textit{what} tools to call, but \textit{why} and \textit{how} to reason between them. Details for our trajectory data curation can be found in Appendix~\ref{apdx:data_curation}.

\subsection{Multi-turn Tool GRPO}

\label{sec:multi-turn-grpo}

To train our model for complex multi-turn tool-planning scenarios, we extend the GRPO framework to effectively handle multi-turn tool-calling reasoning trajectories. Concretely, we use \textbf{Multi-turn Reward Accumulation} and \textbf{Adaptive Tool Reward} to ensure the efficacy of the RL procedure.

\textbf{Multi-turn Reward Accumulation}
Our total reward, $R_{\text{total}}$ is formulated as $R_{\text{total}} = R_{\text{format}} \cdot  (\lambda_{\text{tool}} \cdot  R_{\text{tool}} + \lambda_{\text{acc}} \cdot  R_{\text{acc}})$, with each component adapted for multi-turn trajectories $\tau = \{\tau_0, \dots, \tau_T\}$.
\vspace{-5pt}
\begin{itemize}[leftmargin=*]
    \item \textbf{Format Reward} $R_{\text{format}} = \prod_{i=1}^{n} R_{format}(\tau_i)$ Correct formatting is mandatory at every step. Therefore, the overall format reward for a trajectory is set to 1 if and only if every individual step within it is correctly formatted. A single format error at any turn results in $R_{\text{format}}=0$, nullifying the entire reward for the trajectory. This enforces strict adherence to the reasoning structure.

    \item \textbf{Tool Reward} ~The overall tool reward is the average of the fine-grained scores from all tool-calling turns (from $\tau_0$ to $\tau_{T-1}$). It is calculated as $R_{\text{tool}} = \frac{1}{T} \sum_{t=0}^{T-1} R_{\text{tool}}(\tau_t)$. Each individual tool call, $R_{\text{tool}}(\tau_t)$, is evaluated using a hierarchical score of 0-4 based on four criteria (Structure, Name, Parameter Name, and Parameter Content).

    \item \textbf{Accuracy Reward} This reward is granted only based on the final turn, $\tau_T$. If the final answer is correct, $R_{\text{acc}}=1$; otherwise, it is 0.
\end{itemize}

\textbf{Adaptive Reward for Encouraging Tool Use.}

To encourage reliable tool usage under uncertainty, we design an adaptive reward with an asymmetric structure.
When the final prediction is correct, the model receives a full reward regardless of tool usage, encouraging concise reasoning without unnecessary tool calls. When the prediction is incorrect, the reward depends on the quality of tool usage, granting partial credit to informative tool-based reasoning while penalizing unsupported guessing.
This design treats tools as a fallback mechanism under uncertainty rather than a mandatory step, leading to more robust and adaptive decision-making. Details are provided in Appendix~\ref{apdx:rl}.

\subsection{Adaptive Learning for Improved Generalization}
\label{sec:adaptive_learning}

To bolster the model's generalization capabilities when encountering unseen tools and new tasks, we introduce an \textbf{Adaptive Learning} strategy. 
Concretely, we randomize tool names as well as their parameter names and descriptions, preventing the model from overfitting to fixed identifiers and encouraging it to rely on the semantic understanding of tool functionalities and parameter descriptions when selecting appropriate tools.

\noindent\textbf{Token-Level Randomization for Identifiers.} 
We hypothesize that a robust tool planner should not depend on semantically meaningful identifiers (e.g., relying on the word ``Calculator'' to know it performs math). To test and enforce this, we apply a replacement policy to all functional identifiers, including tool names and argument names.
Specifically, these identifiers are replaced with completely random alphanumeric strings (e.g., replacing \texttt{GetWeather} with \texttt{Func\_X7a2}). This process strips away the linguistic cues from the identifiers, compelling the model to infer functionality solely from the provided descriptions and context.

\noindent\textbf{Semantic-Level Paraphrasing for Descriptions.} 
While identifiers are randomized, it remains crucial to maintain sufficient diversity in the descriptions while preserving their semantic integrity to ensure learnability.  Therefore, for the descriptions of both tools and arguments, we employ a semantic rephrasing strategy.
We leverage Gemini 2.5 Flash to rephrase the original descriptions.
The objective is to alter the syntactic structure and lexical choices while strictly maintaining the original functional meaning. This creates a diverse set of equivalent tool definitions, preventing the model from overfitting to specific phrasing and enhancing its robustness to the variations in tool documentation encountered in real-world scenarios.

\section{Experiments}

In this section, we describe our experimental setup in detail, analyze the impact of tool augmentation on agent planning, investigate the adaptability and generalization of learned tool-using behaviors, and finally evaluate the resulting model across diverse multimodal reasoning tasks.

Our core experiments are based on the Qwen2.5-VL-3B-Instruct and Qwen2.5-VL-7B-Instruct models \citep{qwen25vl}. 
We evaluate our approach across four diverse tasks that probe complementary aspects of multimodal reasoning.
Specifically, we consider \textbf{Visual Spatial Planning}, which assesses multi-step planning and perceptual grounding, evaluated on our custom out-of-distribution benchmark (VSPO) as well as the standard VSP benchmark~\citep{wu2024vsppaper}.
We also include \textbf{Jigsaw}, which focuses on visual compositionality, evaluated on both our Jigsaw-COCO dataset and the Jigsaw subset of BLINK~\citep{fu2024blink}.
To assess fine-grained visual understanding of GUI scenarios, we further evaluate on \textbf{GUIQA}, using GUIChat~\citep{chen2024guicourse} and WebQA from the WebMMU benchmark~\citep{awal2025webmmu}. To better evaluate the effectiveness of our visual tools in GUI-based scenarios, we focus on the agent acting subset of the English split of WebMMU. Detailed settings and implementation details for all tasks are provided in Appendix~\ref{apdx:task_definition}.

\subsection{Visual Tools Help Bridging the Capability Gap}
\label{sec:exp_single_task_train}

We first fine-tune the model on each individual task and systematically evaluate the contribution of TC and TG to the overall performance.
We include Direct SFT and Direct GRPO as strong baseline settings.
Specifically, Direct SFT refers to supervised fine-tuning of the base model on the training set of each individual task, while Direct GRPO applies rule-based GRPO to the base model to enhance its reasoning capability following prior work~\citep{r1zero_aha_moment}.
The results are summarized in Table~\ref{tab:main_table}.

\begin{table*}[h]
\centering
\small
\tablestyle{11pt}{1.0}
\resizebox{\textwidth}{!}{%
\begin{tabular}{l|c|c|c|c|c|c}
\toprule
\textbf{Model} & 
\textbf{VSPO} & 
\textbf{VSP} & 
\textbf{Jigsaw} & 
\textbf{BLINK-J} & 
\textbf{GUIChat} & 
\textbf{WebMMU$^\dagger$} \\
\midrule
Qwen2.5-VL 3B 
& 26.53 & 26.73 & 39.80 & 48.67 & 45.11 & 55.89 \\

\;\;+ Direct SFT 
& 38.15 & 38.82 & 42.60 & 53.33 & 55.51 & 61.38 \\

\;\;+ Direct GRPO 
& 24.55 & 32.73 & 42.70 & 52.67 & 52.49 & 56.30 \\

\;\;+ TC (cold-start) 
& 47.01 & 51.09 & 66.00 & 70.00 & 45.32 & 44.72 \\

\;\;+ TG (tool-GRPO) 
& 29.10 & 35.09 & 43.00 & 47.33 & 89.60 & \underline{72.15} \\

\rowcolor{gray!12}
\;\;\textbf{+ TC + TG} 
& \underline{84.73} & \underline{94.73} & \underline{94.80} & \underline{88.67} & 85.45 & \textbf{81.71} \\

\rowcolor{gray!6}
\textbf{$\Delta$ vs.\ base} 
& {\color{ForestGreen}+58.20} 
& {\color{ForestGreen}+68.00} 
& {\color{ForestGreen}+55.00} 
& {\color{ForestGreen}+40.00} 
& {\color{ForestGreen}+40.34} 
& {\color{ForestGreen}+25.82} \\
\midrule

Qwen2.5-VL 7B 
& 25.39 & 28.09 & 45.70 & 52.67 & 59.46 & 67.48 \\

\;\;+ Direct SFT 
& 42.18 & 46.64 & \underline{86.40} & \underline{88.00} & 62.68 & 65.65 \\

\;\;+ Direct GRPO 
& 28.38 & 30.18 & 64.90 & 80.00 & 67.67 & \underline{83.54} \\

\;\;+ TC (cold-start) 
& \underline{61.69} & \underline{64.91} & 84.20 & 83.33 & 61.85 & 64.63 \\

\;\;+ TG (tool-GRPO) 
& 59.70 & 73.18 & 72.30 & 80.67 & \textbf{92.52} & \textbf{88.62} \\

\rowcolor{gray!12}
\;\;\textbf{+ TC + TG} 
& \textbf{85.09} & \textbf{97.64} & \textbf{96.60} & \textbf{96.00} & \underline{88.57} & 82.32 \\

\rowcolor{gray!6}
\textbf{$\Delta$ vs.\ base} 
& {\color{ForestGreen}+59.70} 
& {\color{ForestGreen}+69.55} 
& {\color{ForestGreen}+50.90} 
& {\color{ForestGreen}+43.33} 
& {\color{ForestGreen}+29.11} 
& {\color{ForestGreen}+14.84} \\
\bottomrule
\end{tabular}
}
\caption{Comparison of the Effects of Tool Cold-Start (TC) and Tool-GRPO (TG) under a Single-Task Fine-Tuning Setting.
$^\dagger$WebMMU reports the Agentic Action (Act.) score.
Best is \textbf{bold}, second-best is \underline{underlined}.
Detailed sub-task breakdown is provided in Appendix Table~\ref{tab:full_results}.}
\label{tab:main_table}
\end{table*}

\textbf{Visual Tools Bring Stable Improvements}~~~ As shown in Table~\ref{tab:main_table}, our TC + TG recipe consistently improves the performance of base models, demonstrating an average gain of \textcolor{ForestGreen}{\textbf{+38.66\%}} on the 7B model. This tool-augmented approach transforms tasks like VSP from an under-optimized baseline ($\sim$31.64\%) to near-perfect execution (\textbf{97.64\%}). This performance significantly surpasses traditional optimization methods such as task-specific SFT (\textbf{46.64\%}) and Direct GRPO (\textbf{30.18\%}). Furthermore, our TC + TG recipe enables the 7B model to achieve the state-of-the-art results that are competitive with, or superior to the best proprietary models. For instance, on VSP and Jigsaw, our model outperforms GPT-5 \citep{openai2025gpt5} (\textbf{96.60\% vs. 80.10\%}). This confirms that our method is a highly effective strategy for unlocking advanced reasoning capabilities in open-source models.

\textbf{Visual Tools Help Overcome Scale-Based Limitations} ~~~The results also reveal that tool augmentation can redefine the performance ceiling of MLLMs by overcoming scale-based limitations. As illustrated in Table~\ref{tab:main_table} and Figure~\ref{fig:tool_model_boundary}, while the baseline performance of 3B and 7B models is disparate and low, our tool-augmented versions both achieve near-perfect accuracy (94.7\% and 97.6\%). This suggests that the primary performance bottleneck has shifted from the model’s scale to the intrinsic quality of the tools it employs.

\textbf{Why Visual Tools Help}~~~ Our empirical analysis reveals that the effectiveness of visual tools stems from their complementary roles in enhancing \emph{perception}, \emph{verification}, and \emph{planning}. First, expert perception tools compensate for the intrinsic visual limitations of MLLMs by providing precise, structured observations, substantially improving downstream reasoning even in zero-shot settings. Second, manipulation tools enable models to externalize intermediate hypotheses and verify them through explicit visual operations, transforming abstract reasoning into concrete, checkable decisions. Finally, high-quality tool-augmented trajectories further teach models when and how to apply these tools, aligning planning behavior with task structure. Together, these components shift the bottleneck from internal reasoning accuracy to effective tool planning, explaining the consistent and large performance gains observed across tasks. Detailed analyses are provided in Appendix \ref{apdx:why_visual_tools_help}.

\subsection{\method~ Can Learn Adaptive Tool-Using}
\label{sec:exp_adaptive}
To investigate whether MLLMs can effectively learn to select tools and adaptively regulate their usage frequency, we carried out a systematic study to build an adaptive tool planning model, which is the main characteristic of our \method. We choose VSP as the primary testbed for this adaptive study, as it comprises multiple interdependent subtasks (e.g., navigation and verification) and requires more complex tool planning decisions involving diverse tool types. 

\begin{table}[h]
\centering
\small  % 添加这行，在 resizebox 之前设置基础字号
\tablestyle{7pt}{1.1}
\resizebox{\textwidth}{!}{%
\begin{tabular}{lcccccccccc}
\toprule
\multirow{2}{*}{\textbf{Stage}} & 
\multirow{2}{*}{\textbf{Reflection}} & 
\multirow{2}{*}{\textbf{A*}} & 
\multicolumn{3}{c}{\textbf{VSP}} &  % 添加加粗
\multicolumn{3}{c}{\textbf{VSPO}} &  % 添加加粗
\multicolumn{2}{c}{\textbf{A* Statistics}} \\  % 添加加粗
\cmidrule(lr){4-6} \cmidrule(lr){7-9} \cmidrule(lr){10-11}
& & & Nav & Verify & Overall & Nav & Verify & Overall & CPS & Succ. \\  % Succ Rate → Succ. 省空间
\midrule
\rowcolor{gray!8}  % 轻微高亮第一行（完整方法）
TC + TG & \ding{55} & RL & \textbf{96.33} & \underline{99.20} & \textbf{97.64} & \textbf{73.44} & \underline{98.70} & \textbf{85.09} & 0.56 & \textbf{100.0} \\  % 统一小数位
TC + TG & \checkmark & - & \underline{84.33} & \textbf{99.80} & \underline{91.36} & \underline{63.89} & \textbf{99.61} & \underline{80.36} & 0.00 & 0.0 \\
TC + TG & \checkmark & Inf & 55.17 & 84.60 & 68.55 & 57.22 & \textbf{99.61} & 76.77 & \underline{0.68} & 16.9 \\
TC + TG & \ding{55} & Inf & 62.33 & 80.00 & 70.36 & 43.78 & 88.70 & 64.49 & 0.52 & \underline{94.5} \\
TC Only & \ding{55} & - & 41.00 & 93.60 & 64.91 & 31.58 & 94.01 & 61.69 & 0.00 & 0.0 \\
TC + TG & \ding{55} & - & 44.83 & 94.20 & 67.27 & 27.67 & 94.81 & 58.62 & 0.00 & 0.0 \\
TC Only & \ding{55} & Inf & 46.00 & 79.40 & 61.18 & 32.11 & 81.43 & 54.85 & \textbf{0.49} & 85.2 \\
\bottomrule
\end{tabular}
}

\caption{\textbf{Ablation study on adaptive tool usage.}  % 短标题加粗
\textbf{Stage} compares Tool Cold Start (TC) + Tool GRPO (TG) against TC alone. 
\textbf{Reflection} indicates training with (\checkmark) or without (\ding{55}) reflection data. 
\textbf{A*} specifies availability: during RL training, at Inference (Inf), or unavailable (-). 
\textbf{A* Statistics:} CPS = calls per sample, Succ. = success rate (\%). 
Best is \textbf{bold}, second-best is \underline{underlined}.}
\label{tab:adaptive_study}
\end{table}

\paragraph{\method~ Can Use New Tools during Inference Time} 
During inference, the model exhibits a notable ability to adapt to previously unseen but powerful tools.
To systematically examine this capability, we evaluate whether the model can effectively leverage a strong planning tool, \textsc{A*}, which is deliberately excluded during the Tool Cold Start (TC) phase.
As shown in Table \ref{tab:adaptive_study}, when the A* tool is introduced solely at inference time (Inf), it provides a significant performance boost to the relevant task. For our standard TC+TG model (without reflection), this elevates the VSP navigation score from 44.83 to 62.33. The A* Statistics corroborate this adaptive behavior, showing a high invocation success rate of \textbf{94.53\%}, which indicates the model is not just guessing but is correctly learning the tool's syntax and purpose in a zero-shot setting.

However, although the model exhibits adaptive tool invocation behavior at inference time, this adaptability remains unstable and inconsistent.
For example, the \textsc{Astar} tool does not contribute to verification and instead acts as a distractor in this task. When the tool is made available, the model nevertheless invokes it, leading to a substantial performance drop from 94.20 to 80.00. Similar unstable tool invocation behaviors become more pronounced when reflection is enabled.
Therefore, the adaptability to new tools during inference remains unstable and still requires RL to stabilize it.

\begin{figure}[h!]
    \centering

    \begin{subfigure}{0.32\textwidth}
        \centering
        \includegraphics[width=\linewidth]{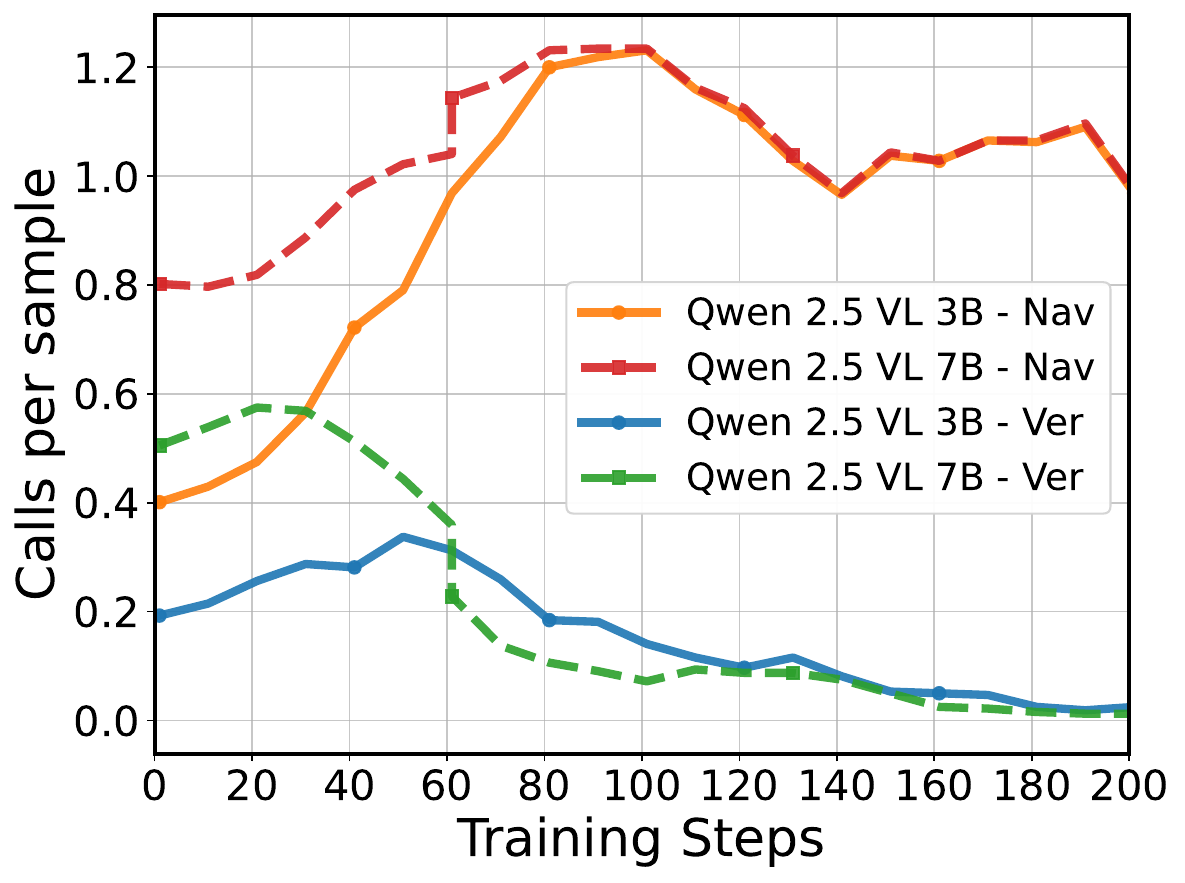}
        \caption{\textsc{AStar}}
        \label{fig:toolcall_astar}
    \end{subfigure}
    \hfill
     \begin{subfigure}{0.32\textwidth}
        \centering
        \includegraphics[width=\linewidth]{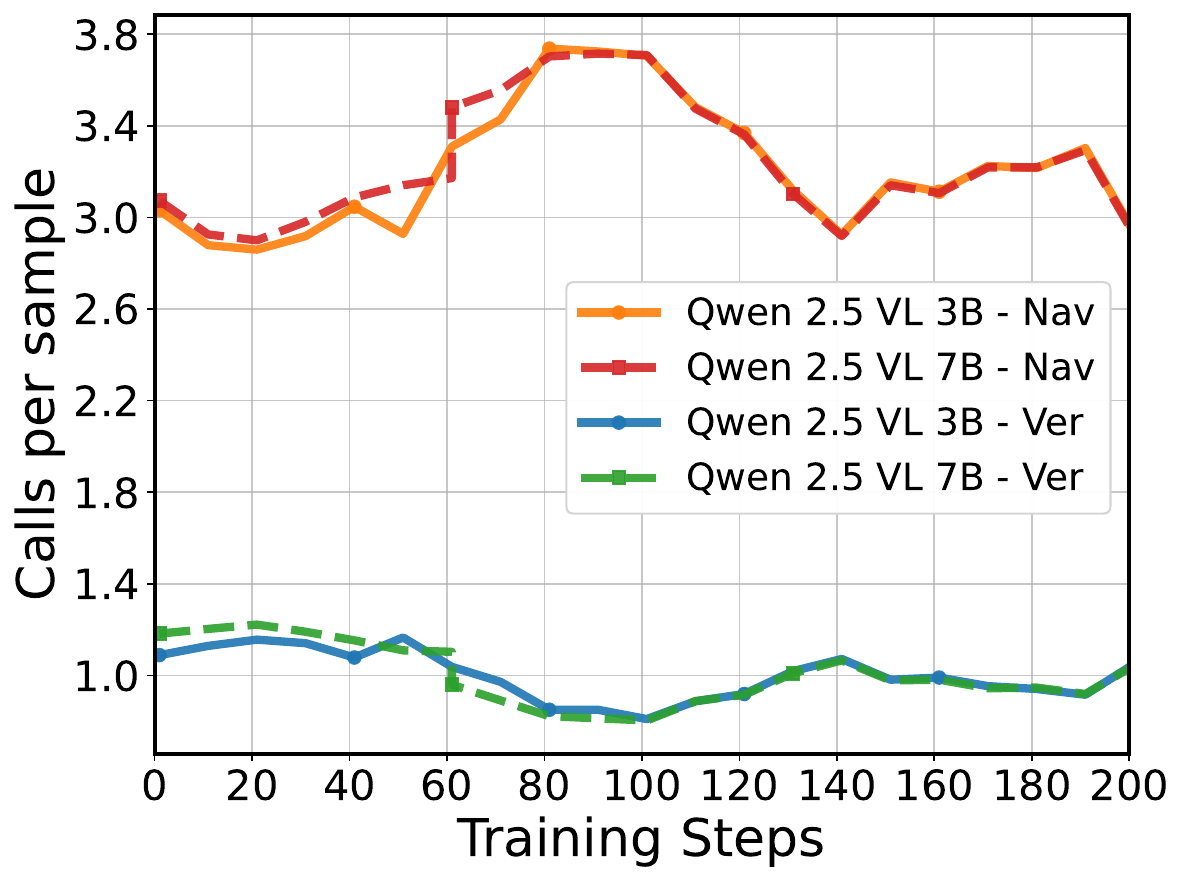}
        \caption{\textsc{Point}}
        \label{fig:toolcall_point}
    \end{subfigure}
    \hfill
    \begin{subfigure}{0.32\textwidth}
        \centering
        \includegraphics[width=\linewidth]{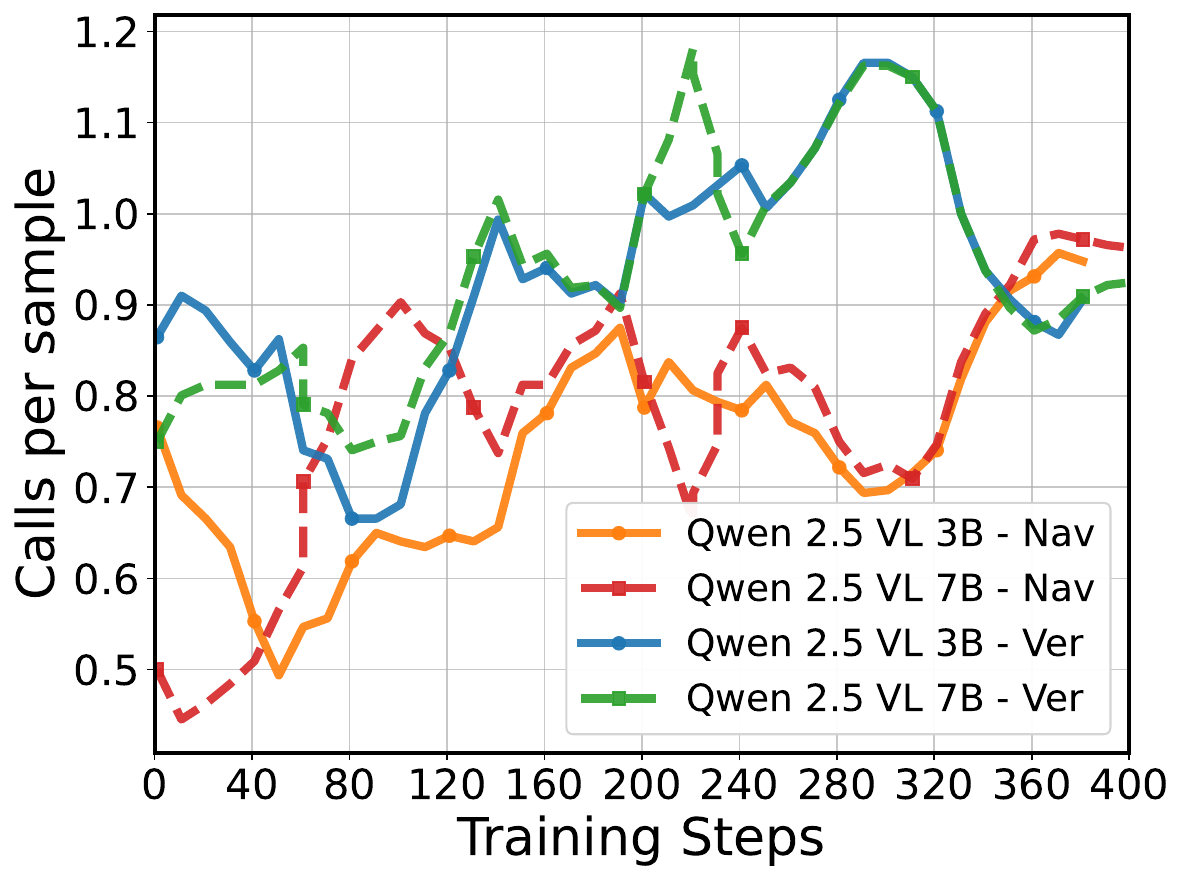}
        \caption{\textsc{Draw2DPath}}
        \label{fig:toolcall_draw2dpath}
    \end{subfigure}
    \vspace{10pt}
    \caption{Trend for tool calling frequencies for \textsc{AStar}, \textsc{Point}, and \textsc{Draw2DPath} during RL. The model is optimized on VSP Verification (cool-color) and VSP Navigation (warm-color) tasks.}
    \label{fig:tool_call_frequency}
\end{figure}

\paragraph{Learning to Adopt and Master Beneficial Tools through RL} 
To improve the stability of adaptive tool invocation, we incorporate the \textsc{A*} tool into the tool set during the TG stage. Specifically, we initialize training from the same SFT checkpoint that has never been exposed to \textsc{A*}, and then introduce \textsc{A*} as an available tool during the TG process. As shown in Table~\ref{tab:adaptive_study}, incorporating \textsc{A*} during RL leads to more stable tool usage patterns and yields consistently improved performance.
In addition, we also observe three notable adaptive behaviors emerging during the TG stage.

\begin{enumerate}[leftmargin=*, label=(\arabic*)]
\item \textbf{Learning to Adopt Beneficial Tools.} As illustrated in Figure \ref{fig:toolcall_astar}, for the Path Navigation task (warm-colored curves), the model's invocation frequency for \textsc{AStar} progressively increases, stabilizing at a high rate of over \emph{1.0 call} per sample. This upward trajectory indicates that the model, guided by task-completion rewards, correctly identifies \textsc{AStar} as a highly beneficial tool for pathfinding and actively incorporates it into its problem-solving strategy. As a result, this mastery translates to a dramatic performance increase, with the VSP navigation score soaring to \textbf{96.33}, which achieves the best performance.

\item \textbf{Learning to Discard Irrelevant Tools.}
Critically, \method~learns to discard the tool when it is irrelevant. Figure \ref{fig:toolcall_astar} (cool-colored curves) shows the inverse trend for the Verification task. The model initially explores using the A* tool but, receiving no reward for doing so, gradually learns to suppress its usage, with the invocation frequency decaying towards \emph{zero}. This adaptive pruning prevents the negative degradation observed in the zero-shot inference scenario, allowing the Verification performance to remain at a near-perfect 99.20.

\item \textbf{Learning to Modulate Tool-Use Frequency}
Beyond the binary choice of adopting or discarding a tool, the model exhibits a more nuanced adaptive behavior: dynamically modulating the invocation frequency of continuously useful tools to find an optimal balance. 
For instance, as shown in Figures \ref{fig:toolcall_point} and \ref{fig:toolcall_draw2dpath}, the model learns that the Point tool is significantly more critical for navigation, maintaining a high and stable call frequency ($\sim3.2$ calls/sample), while keeping its usage minimal for verification ($\sim1.0$ call/sample). 

\end{enumerate}

\begin{table*}[t]
\centering
\small
\tablestyle{8pt}{1.1}
\resizebox{\textwidth}{!}{
\begin{tabular}{l|c|c|c|c|c|c}
\toprule
\textbf{Model} & 
\textbf{VSPO} & 
\textbf{VSP} & 
\textbf{Jigsaw} & 
\textbf{BLINK-J} & 
\textbf{GUIChat} & 
\textbf{WebMMU}$^\dagger$ \\
\midrule
Qwen2.5-VL 7B 
& 25.39 & 28.09 & 45.70 & 52.67 & 68.09 & 67.48 \\

\;\;+ TC 
& 27.66 & 29.36 & \underline{83.50} & \underline{82.00} & 36.59 & 41.87 \\

\;\;+ Rnd TC 
& 26.71 & 30.36 & 77.90 & 80.00 & 35.55 & 42.89 \\

\;\;+ TG 
& 24.01 & 30.73 & 66.10 & 82.00 & \underline{82.54} & 67.68 \\

\;\;+ TC + TG 
& 24.43 & 28.00 & 59.20 & 78.00 & 80.35 & 59.76 \\

\;\;+ Rnd TC + TG 
& \underline{57.78} & \underline{47.27} & \textbf{90.60} & \textbf{91.33} & \textbf{82.74} & \underline{69.11} \\

\rowcolor{gray!12}
\;\;\textbf{+ Rnd TC + Rnd TG (Ours)} 
& \textbf{69.94} & \textbf{78.91} & 83.50 & 80.67 & 80.87 & \textbf{70.93} \\

\rowcolor{gray!6}
\textbf{$\Delta$ vs.\ base} 
& {\color{ForestGreen}+44.55} & 
{\color{ForestGreen}+50.82} & 
{\color{ForestGreen}+37.80} & 
{\color{ForestGreen}+28.00} & 
{\color{ForestGreen}+12.79} & 
{\color{ForestGreen}+3.46} \\
\bottomrule
\end{tabular}
}
\caption{Impact of Different Experimental Settings on Model Generalization Performance.
\textbf{Rnd TC} and \textbf{Rnd TG} denote the randomized Tool Cold Start and Tool GRPO settings trained using our Adaptive Learning Method.
$^\dagger$\textbf{WebMMU} reports the Agentic Action (Act.) performance.
Best is \textbf{bold}, second-best is \underline{underlined}.
Full sub-task breakdown in Appendix Table~\ref{tab:generalization_full}.}
\label{tab:generalization_1_to_3}
\end{table*}

\subsection{\method~ Can Learn Generalized Tool-Using}
\label{sec:generalization_1_to_3}

While adaptability to new tools within a known context is valuable, a more fundamental challenge lies in \textbf{cross-tool and cross-task generalization}.
In practice, simply applying our TC + TG training recipe is not sufficient to achieve robust generalization: models may still fail when confronted with previously \emph{unseen tools} or \emph{novel task distributions}. To overcome this limitation, 
we propose \textit{Adaptive Learning} method that randomizes tool definitions at both the token level and the semantic level. This strategy can be integrated into both the TC and TG stages. We evaluate its effectiveness from two complementary perspectives:

\textbf{Generalize to New Tasks} ~~ To validate whether the tool-planning capability can generalize to new tasks, we conduct TC only using the \textbf{Jigsaw} data, while withholding all training data related to the VSP and WebQA tasks.  We then apply various training recipes to investigate whether tool-planning capabilities acquired from Jigsaw trajectories could be effectively transferred to these unseen domains. All data from the three tasks is used during the TG stage.

\textbf{Generalize to New Tools} ~~ To validate whether the tool-planning capability can generalize to new tools, the toolset definition during evaluation is \textbf{completely different} from the one used during training, while preserving the same underlying tool functionalities. This design allows us to assess the model’s ability to generalize to unseen tools that provide equivalent functionality, thereby evaluating its robustness and generalizability beyond memorizing tool-specific interfaces.

The results, as shown in Table \ref{tab:generalization_1_to_3}, reveal the potential for generalized tool-using, demonstrating that tool-planning skills learned from a single task can generalize to unseen domains.
Specifically, the model trained with Randomized TC + Randomized TG achieves the most significant improvement over the base model. On the unseen VSP task, it boosts the overall score from \textbf{28.09} to \textbf{78.91}, and also demonstrates strong generalization when averaged across all three tasks (\textbf{75.81 vs. 46.50}). In contrast, other training settings fail to improve performance on unseen tasks and, in some cases, even degrade it. These results indicate that our method enables the model to learn generalizable tool usage, rather than overfitting to task-specific patterns.

\begin{table*}[t]
\centering
\small
\tablestyle{1.5pt}{1.1}
\resizebox{\textwidth}{!}{
\begin{tabular}{lcccccccc|c}
\toprule
\textbf{Model} & 
\textbf{~~VSPO~~} & 
\textbf{~~VSP~~}& 
\textbf{~Jigsaw~} & 
\textbf{BLINK-J} & 
\textbf{GUIChat} & 
\textbf{WebMMU} & 
\textbf{HRBench} & 
\textbf{~~~~V*~~~~} &
\textbf{~~~~Avg.~~~~} \\
\midrule
\multicolumn{10}{c}{\textit{\textbf{Closed-Source Models}}} \\
\midrule
Gemini 2.5 flash & 40.12 & 53.55 & 67.20 & 65.33 & 83.05 & 66.26 & 78.25 & 68.59 & 65.54 \\
GPT 5 & 34.25 & 55.64 & 80.10 & 73.33 & 71.41 & 80.49 & 74.38 & 74.87 & 68.56 \\
Claude 4 sonnet & 51.56 & 56.27 & 58.60 & 65.33 & 93.14 & 83.54 & 60.62 & 59.69 & 66.09 \\
\midrule
\multicolumn{10}{c}{\textit{\textbf{Open-Source Models}}} \\
\midrule
Qwen 2.5 VL 3B & 26.53 & 26.73 & 39.80 & 48.67 & 46.26 & 54.47 & 53.00 & 43.98 & 42.93 \\
Qwen 2.5 VL 7B & 25.39 & 28.09 & 45.70 & 52.67 & 68.09 & 67.48 & 63.62 & 63.35 & 51.80 \\
Qwen 2.5 VL 32B & 28.56 & 33.91 & 59.50 & 64.67 & 85.21 & 85.98 & 70.12 & 72.25 & 62.53 \\
Qwen 2.5 VL 72B & 33.41 & 39.09 & 70.10 & 71.33 & 88.01 & 91.06 & 73.00 & 80.10 & 68.76 \\
InternVL3 78B & 28.14 & 35.09 & 52.80 & 60.00 & 79.83 & 71.34 & 75.12 & 81.15 & 60.93 \\
\midrule
\multicolumn{10}{c}{\textit{\textbf{Tool-Planning Models}}} \\
\midrule
Qwen 2.5 VL 7B + Tools 
& 28.98 & 30.45 & 45.00 & 56.00 & 56.76 & 69.51 & 63.75 & 54.97 & 50.93 \\
Qwen 2.5 VL 72B + Tools 
& 39.64 & 45.00 & 61.50 & 65.33 & \textbf{77.13} & \underline{76.83} & 67.12 & 65.97 & 62.82 \\
GPT 5 + Tools 
& \underline{52.75} & \underline{71.36} & \underline{84.50} & \underline{76.00} & \underline{76.51} & \textbf{88.82} & \textbf{78.50} & \underline{70.16} & \underline{74.83} \\
DeepEyes$^*$ 
& 10.18 & 12.18 & 44.80 & 50.67 & 65.90 & 72.76 & 67.00 & 67.54 & 48.88 \\
PixelReasoner$^*$ 
& 21.38 & 24.55 & 52.20 & 59.33 & 72.45 & 69.51 & 52.12 & 53.40 & 50.87 \\
\rowcolor{gray!12}
\textbf{AdaReasoner 7B} 
& \textbf{71.08} & \textbf{78.36} & \textbf{88.60} & \textbf{88.00} 
& 73.91 & 72.15 & \underline{69.12} & \textbf{70.68} & \textbf{76.49} \\
\rowcolor{gray!6}
\textbf{$\Delta$ vs.\ base} & 
{\color{ForestGreen}+45.69} & 
{\color{ForestGreen}+50.27} & 
{\color{ForestGreen}+42.90} & 
{\color{ForestGreen}+35.33} & 
{\color{ForestGreen}+5.82} & 
{\color{ForestGreen}+4.67} & 
{\color{ForestGreen}+5.50} & 
{\color{ForestGreen}+7.33} & 
{\color{ForestGreen}+24.72} \\
\bottomrule
\end{tabular}
}
\caption{\textbf{Main results on visual reasoning, WebQA, and general VQA tasks.}
$^\dagger$WebMMU reports the Agentic Action (Act.) score.
\textbf{Avg} denotes the average score across all benchmarks.
\textbf{Bold} and \underline{underline} indicate the best and second-best results within tool-planning models. Detailed results are provided in Appendix \ref{apdx:detailed_res} and Appendix Table \ref{tab:part4_full}. $^*$All evaluations are conducted under a unified evaluation framework.}  
\label{tab:part4_main_table}
\end{table*}

\begin{table*}[h]
\centering
\small
\tablestyle{1.5pt}{1.1}
\begin{tabular}{l|cccc|ccccc}
\toprule
\multirow{2}{*}{\textbf{Model}}
& \multicolumn{4}{c|}{\textbf{Jigsaw}}
& \multicolumn{5}{c}{\textbf{Vstar}} \\
\cmidrule(lr){2-5} \cmidrule(lr){6-10}
& \# Turns & CPS & Succ(\%) & ~~Acc.~~
& \# Turns & CPS & Succ(\%) & Acc. & Orig. Acc. \\
\midrule
DeepEyes 
& 1.51 & 1.09 & 22.71 & 44.80
& 1.01 & 0.02 & 25.00 & 67.54 & 90.10 \\
PixelReasoner 
& 1.13 & 0.77 & 2.09 & 52.20
& 1.19 & 0.26 & 32.00 & 53.40 & 84.30 \\
GPT5 + Tools 
& 1.05 & 0.00 & 0.00 & \underline{84.50}
& 1.24 & 0.24 & 89.13 & \underline{70.16} & -- \\
Qwen 2.5 VL 7B + Tools 
& 3.40 & \underline{2.06} & 64.41 & 45.00
& \underline{1.91} & 0.18 & \underline{94.29} & 54.97 & -- \\
Qwen 2.5 VL 72B + Tools 
& \underline{4.24} & 0.57 & \underline{90.40} & 61.50
& 1.72 & \underline{0.27} & \textbf{100.00} & 65.97 & -- \\
\rowcolor{gray!12}
\textbf{AdaReasoner 7B} 
& \textbf{4.49} & \textbf{3.54} & \textbf{98.50} & \textbf{88.60}
& \textbf{2.35} & \textbf{1.47} & 90.04 & \textbf{70.68} & -- \\
\bottomrule
\end{tabular}
\caption{\textbf{Tool usage statistics on Jigsaw and Vstar.}
\textbf{\#Turns} denotes the number of interaction rounds.
\textbf{CPS} (calls per sample) represents the average number of tool invocations per sample.
\textbf{Succ} denotes the tool execution success rate.
\textbf{Acc} is taken from the corresponding benchmark results in Table~\ref{tab:part4_main_table}.
\textbf{Orig. Acc.} denotes the performance under the model’s original tool definition.}
\label{tab:tool_statistics_multi_task}
\end{table*}

\subsection{Main Results}
\label{sec:exp_main_res}
Following our proposed training recipe and detailed analysis, we select the data from VSP, Jigsaw, and WebQA as our cold start data and VSP, Jigsaw, WebQA, and Visual Search as our RL data. The training details can be found in Appendix \ref{apdx:final_model_details}. To enhance the model's generalizable tool-use capabilities, we applied the Adaptive learning method (i.e., Randomized TC + Randomized TG) strategy on Qwen 2.5VL 7B.
We evaluate our method on a diverse set of benchmarks. In addition to the benchmarks introduced earlier, we further include visual search benchmarks to assess the generalizability of our approach beyond structured visual reasoning tasks, specifically V*\citep{wu2024vstar} and HRBench\citep{hrbench}. The comparative results are summarized in Table \ref{tab:part4_main_table}.

\textbf{Overall Performance Across Tasks}~~~As shown in the table, our model achieves consistent and substantial performance improvements across both visual reasoning tasks (e.g., VSP and Jigsaw) and general multimodal tasks (e.g., WebQA and Visual Search). In visual reasoning, our method yields large and stable gains ($\Delta > +40\%$), enabling the 7B model to outperform several competitive closed-source models and effectively narrowing the capability gap between smaller open-source models and larger counterparts. For general tasks, while tool integration cannot fully offset the performance advantages brought by model scaling due to their open-ended nature and the lack of deterministic tools, it still provides meaningful and robust performance boosts, demonstrating the broad effectiveness of visual tool augmentation.

\textbf{Generalizable Tool Using.}~~~A defining characteristic of our approach is its robustness against domain shifts in both tool definitions and task contexts.
During evaluation, the tool definitions differ from those used in training, forming a strict zero-shot tool adaptation scenario for all models.
As shown in Table~\ref{tab:tool_statistics_multi_task}, while prior tool-planning methods exhibit limited tool engagement or low execution reliability under this shift, \textbf{AdaReasoner} consistently demonstrates the highest level of effective tool usage.
In particular, it achieves the largest number of calls per sample (CPS) while maintaining near-perfect execution success (e.g., \textbf{3.54 CPS} with \textbf{98.50\%} success on Jigsaw), indicating that the model has learned transferable tool-planning principles rather than overfitting to specific tool APIs.

Crucially, this generalizable tool usage translates into clear performance gains.
On Jigsaw, AdaReasoner attains the best accuracy (\textbf{88.60}), substantially outperforming prior methods, which exhibit limited effectiveness when confronted with previously unseen tasks and tool definitions.
On VStar, which evaluates real-world general VQA scenarios and does not provide explicit tool-calling supervision during training, our model still actively invokes tools (\textbf{1.47 CPS}) and achieves the highest accuracy (\textbf{70.68}).
In contrast, several existing approaches rely heavily on their original tool definitions and suffer notable performance degradation when evaluated under the new setting, as reflected by the gap between their original and adapted accuracies.
Together, these results demonstrate that our method effectively decouples tool-use logic from task-specific supervision, empowering the model to autonomously formulate tool-augmented strategies for new problems.

\begin{figure}[t!]
    \centering
    \includegraphics[width=1\textwidth]{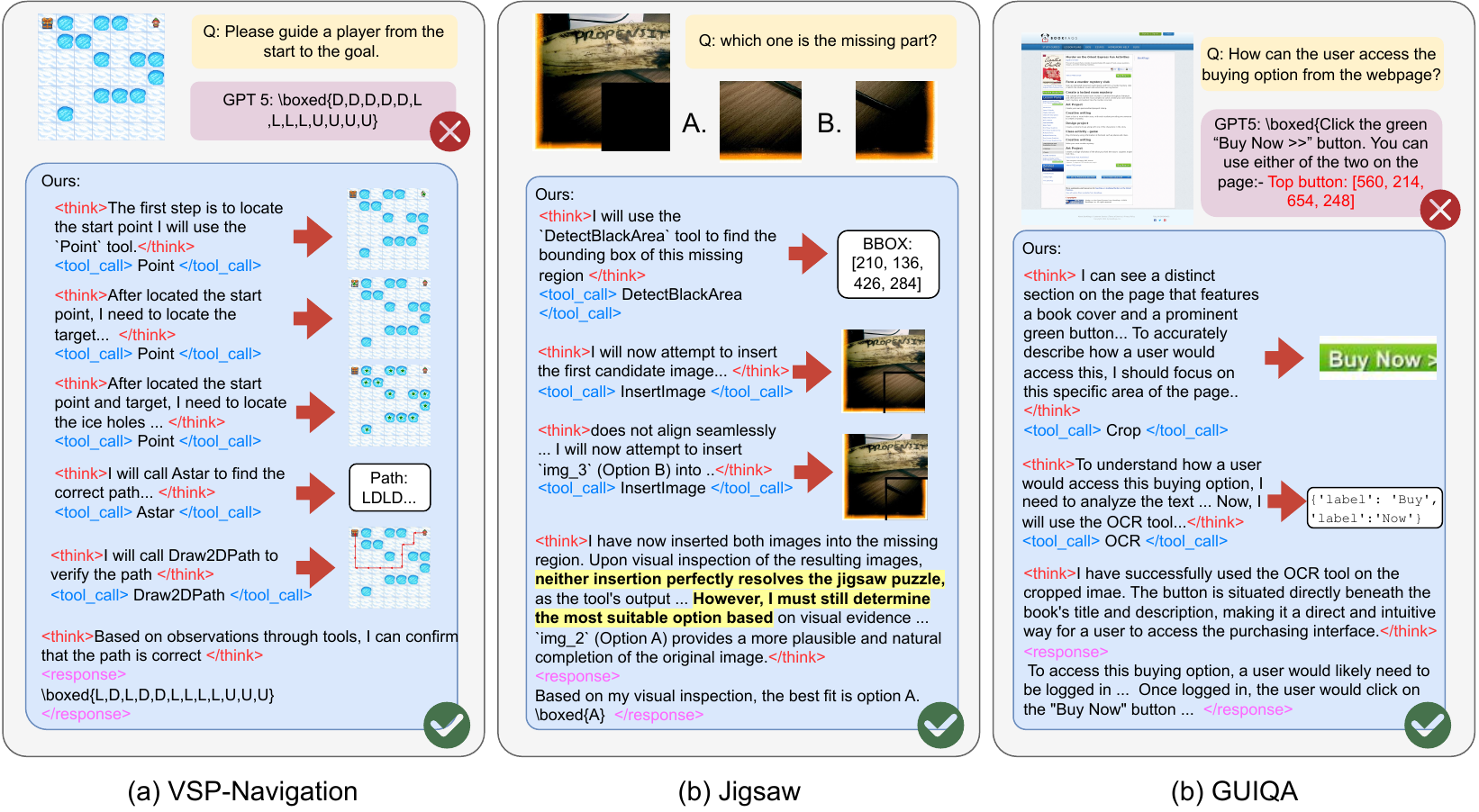}
    \caption{Our \method-7B demonstrates advanced capabilities for multi-turn, tool-assisted reasoning and reflection, enabling it to achieve performance that is on par with, or even superior to, state-of-the-art closed-source models.}
    \label{fig:qualitive_res}
\end{figure}

Moreover, beyond the quantitative improvements reported on benchmark metrics, as shown in Figure~\ref{fig:qualitive_res}, \method-7B exhibits robust qualitative behaviors, including multi-turn tool planning, reflective correction of imperfect tool outputs, and synergistic tool composition, enabling it to outperform strong baselines such as GPT-5.

\section{Related Work}
\subsection{Reinforcement Learning for Multimodal Reasoning}
The recent success of DeepSeek-R1 \citep{guo2025deepseekr1}, which demonstrated that rule-based Group Relative Policy Optimization (GRPO) can effectively induce strong reasoning behaviors in LLMs, has spurred a wave of research aimed at replicating this paradigm in the multimodal domain. Several studies have successfully extended this approach, with \cite{r1zero_aha_moment} reproducing the emergent ``aha'' moment in MLLM reasoning, R1-OneVision \citep{yang2025r1_onevision} introducing a cross-modal formalization pipeline, and works like \cite{feng2025videor1} and \cite{li2025videochatr1} improving temporal reasoning in videos. A collection of other strong works have also leveraged R1-style methods to achieve impressive results in general MLLM reasoning \citep{huang2025visionr1,shen2025vlmr1,lu2025uiR1}. However, a key limitation of the R1-style, rule-based reward structure is that it primarily targets the reasoning process and does not directly improve the model's underlying perceptual abilities. Since accurate perception is the foundation for sound reasoning, error accumulation from faulty perception can still lead to hallucinations and degrade performance. \method~directly addresses this shortcoming. By leveraging the precise perceptual capabilities of external expert models and specialized tools, our framework ensures a high-fidelity understanding of the visual input, thereby improving the reliability of the entire reasoning pipeline.
\subsection{Tool-Augmented Multimodal Reasoning}
There is a growing interest in enhancing MLLMs with tool-use capabilities. Early efforts focused on foundational aspects such as infrastructure and data. LLaVA-Plus \citep{liu2024llavaplus}, for example, introduced a dedicated tool server to provide services for MLLMs. On the data front, CogCoM \citep{qi2024cogcom} identified six key manipulation strategies and trained models on synthetic Chain-of-Manipulation (CoM) data, while TACO \citep{liu2024taco} contributed a large-scale dataset of reasoning traces derived from 15 visual tools.
Subsequent research has explored different paradigms for tool interaction. One prominent line of work enhances visual reasoning by training models to generate code \citep{zhang2025thyme_thinkbyondimage,zhao2507pyvision}. While powerful, these code-based environments are ill-suited for integrating computationally intensive capabilities, such as invoking large expert models. Another line of research leverages simpler, atomic visual tools like zoom-in functions to augment model perception \citep{wang2025vgr,deepeyes,pixelreasoner,zhu2025activeo3,su2025openthinkimg}. However, these approaches typically focus on single-step actions and have not explored the more complex challenges of multi-turn planning or dynamic tool composition. Our work, \method, is designed to bridge these gaps, providing a framework that enables models to perform multi-turn planning and reasoning while adaptively selecting from a diverse suite of tools.

\section{Conclusion}

We introduce a comprehensive framework that integrates high-quality trajectory curation, Tool-GRPO, and an adaptive learning mechanism to enable effective multi-turn tool planning. Based on this framework, we develop AdaReasoner, a family of tool-planning models. Unlike approaches that rely on rigid, task-specific patterns, AdaReasoner learns tool usage as a generalizable reasoning capability, allowing it to coordinate complex tool sequences and adapt zero-shot to unseen tool definitions. Extensive experiments demonstrate that this paradigm achieves state-of-the-art performance, with models exhibiting autonomous adaptive behaviors, selectively adopting beneficial tools, suppressing irrelevant ones, and dynamically modulating tool usage frequency according to task demands. More fundamentally, our findings show that effective tool orchestration shifts the primary performance bottleneck from intrinsic model scale to tool utility, enabling a 7B model to surpass strong proprietary systems such as GPT-5 on challenging visual reasoning tasks.

\renewcommand{\emph}[1]{\textit{#1}}
\bibliography{iclr2026_conference}
\bibliographystyle{iclr2026_conference}

\appendix

\clearpage

\section{Method Details}
\label{apdx:method_details}

\subsection{Basic Settings}
\label{apdx:preliminary}
We first formalize multimodal reasoning with tools as an agentic planning process, enabling a systematic description of how models decompose and solve complex tasks.

\paragraph{Problem Formulation}  We denote an MLLM with tool-using capability as a policy $\pi_\theta$, parameterized by model weights $\theta$. At initialization, $\pi_\theta$ is equipped with access to a pool of tools $\mathcal{T} = {t_1, t_2, \cdots, t_n}$, where $n$ denotes the number of available tools. Given a task description $g$ and the original multimodal input $x = {\text{text}, \text{image}}$, the system begins from an initial state $s_0$.
Building on this setup, the planning framework is formalized through three essential components. \textbf{State} $s_t$ represents the current problem status: the initial state $s_0$ corresponds to the original input, while intermediate states capture textual reasoning steps conditioned on accumulated observations, until a special token triggers an action. \textbf{Action} $a_t$ denotes a one-step tool invocation, delimited by the special symbols \tooltoken{<tool\_call>} and \tooltoken{</tool\_call>}, which executes a selected tool. \textbf{Observation} $o_t$ is the execution result returned by the invoked tool and is incorporated into the subsequent state.

A typical tool-integrated reasoning trajectory $\tau$ involves multiple tool invocations over several reasoning steps, which can be represented as a sequence of rounds:
$$\tau = \{ \tau_0, \tau_1, \tau_2, \dots, \tau_T \}$$
where each round is defined as $\tau_i = \{s_i, a_i, o_i\}$, and the sequence proceeds as follows:
$$s_0 \xrightarrow{a_0} s_1 \xrightarrow{a_1} s_2 \xrightarrow{a_2} \dots \xrightarrow{a_T} s_{T+1}$$

To enable the model to autonomously generate reasoning traces and tool calls, we utilize a system prompt as shown in \ref{apdx:system_prompt} during rollout. The tool list placeholder denotes the tool set $\mathcal{T}$, which contains all tools available for invocation.

\paragraph{Tool Definition and Usage}
\label{apdx:tool-definition}

This section provides a detailed description of the visual tools integrated within our \method~framework. For each tool, we outline its core functionality, input arguments, output format, and its specific role in addressing the challenges of our evaluation tasks.

\begin{itemize}[leftmargin=*,itemsep=5pt]

    \item \textbf{\textsc{Point}}
    \begin{itemize}
        \item \textbf{Functionality:} A perceptual tool designed for precise object localization. Given an image and a natural language description of a target (e.g., ``the start point'', ``red cars''), it returns a list of pixel coordinates (x, y) of the objects' center.
        \item \textbf{Role in VSP:} This tool is fundamental for grounding the model in the spatial environment. In both Navigation and Verification, it is the first step to accurately identify the locations of the start, goal, and hazardous ice holes, converting the visual grid into a structured representation that can be used for planning.
    \end{itemize}

    \item \textbf{\textsc{Draw2DPath}}
    \begin{itemize}
        \item \textbf{Functionality:} A visualization and verification tool. It takes a starting coordinate and a sequence of directional commands (e.g., [`U', `U', `R']) and overlays the corresponding path onto the input image.
        \item \textbf{Role in VSP:} This tool externalizes the model's internal plan into a visual artifact. In Verification, it renders the given path for the model to judge. In Navigation, it serves as a final check, allowing the model to visually confirm that its generated path is correct and safe before outputting the final answer.
    \end{itemize}
    
    \item \textbf{\textsc{AStar}}
    \begin{itemize}
        \item \textbf{Functionality:} A classic planning algorithm encapsulated as a tool. It computes the shortest obstacle-free path between a start and a goal coordinate, given the locations of obstacles.
        \item \textbf{Role in VSP:} This tool offloads the complex pathfinding computation. After the \textsc{Point} tool identifies all key locations, \textsc{AStar} can be invoked to generate an optimal, logically sound path, freeing the MLLM to focus on higher-level task management and verification.
    \end{itemize}
    
    \item \textbf{\textsc{DetectBlackArea}}
    \begin{itemize}
        \item \textbf{Functionality:} A specialized perception tool for the Jigsaw task. It analyzes an image and returns the bounding box coordinates of any completely black, rectangular regions, which correspond to the missing puzzle piece.
        \item \textbf{Role in Jigsaw:} This tool provides a deterministic way to identify the ``problem space''. It is the critical first step in the solution trajectory, telling the model precisely where the candidate patches need to be inserted.
    \end{itemize}

    \item \textbf{\textsc{InsertImage}}
    \begin{itemize}
        \item \textbf{Functionality:} A visual manipulation tool. It takes a base image, a patch image, and a set of coordinates, and returns a new image where the patch has been inserted at the specified location.
        \item \textbf{Role in Jigsaw:} This tool enables iterative hypothesis testing. The model uses it to physically place each candidate patch into the missing slot identified by \textsc{DetectBlackArea}. The resulting composite image is then fed back to the model, allowing it to visually assess the quality of the fit.
    \end{itemize}

    \item \textbf{\textsc{Crop}}
    \begin{itemize}
        \item \textbf{Functionality:} An attentional tool. It takes an image and bounding box coordinates and returns a new, smaller image containing only the specified region.
        \item \textbf{Role in GuiQA:} This tool mimics the human ability to focus on a specific part of a dense interface. By cropping a region of interest (e.g., a button or a text block), the model can reduce noise and ambiguity, creating a cleaner input for subsequent analysis by other tools or by the model itself.
    \end{itemize}

    \item \textbf{\textsc{OCR}}
    \begin{itemize}
        \item \textbf{Functionality:} An information extraction tool. It performs Optical Character Recognition on an input image, returning all detected text along with its corresponding bounding box coordinates.
        \item \textbf{Role in GUIQA:} This tool grounds the model's understanding in the literal text present in the GUI. It is often used synergistically with \textsc{Crop}. After isolating a key UI element, the model can use \textsc{OCR} to reliably read its label (e.g., ``Buy Now''), preventing the visual-only hallucinations common in MLLMs.
    \end{itemize}
\end{itemize}

\begin{figure}[t!]
    \centering
    \includegraphics[width=1\textwidth]{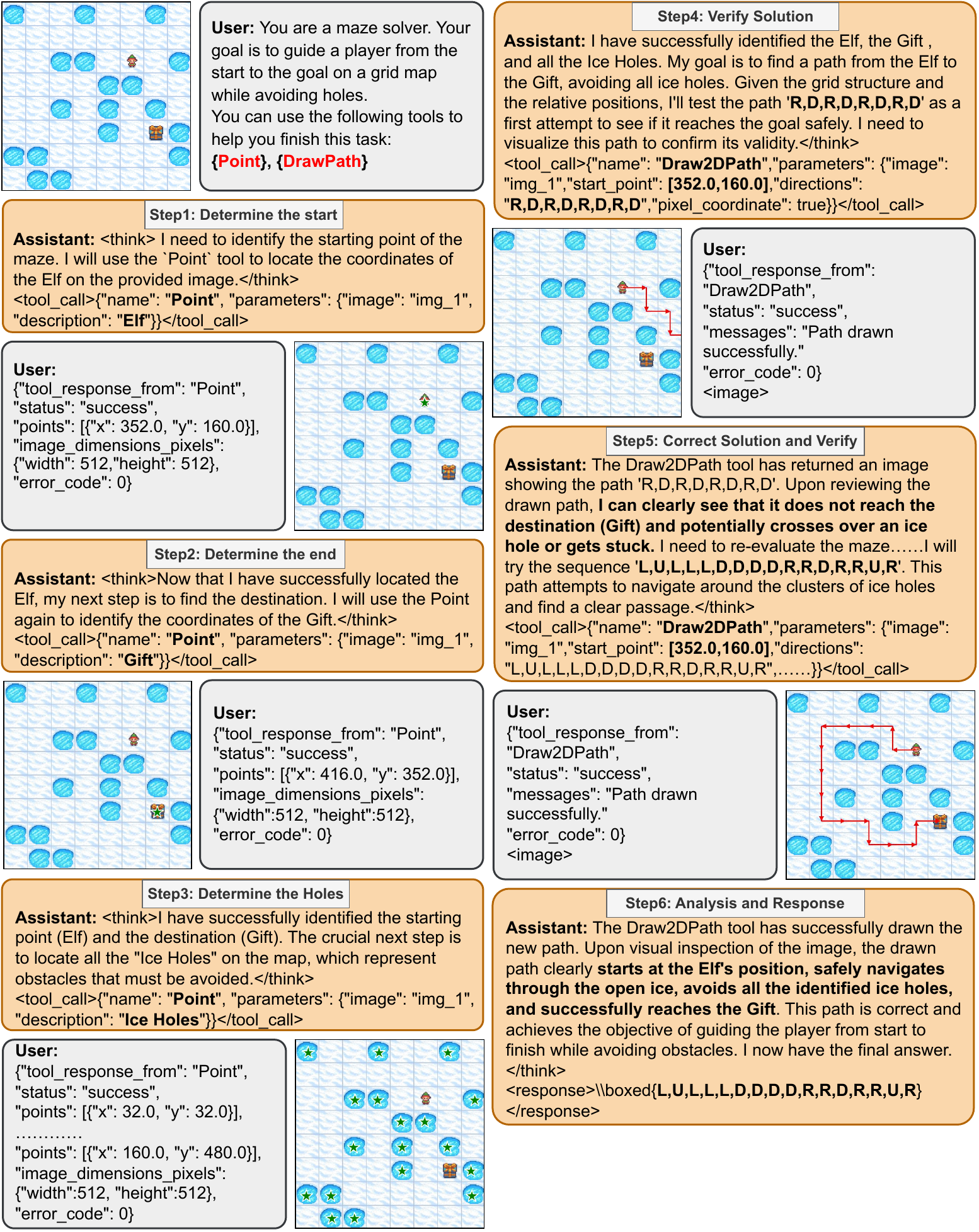}
    \caption{An example of a multi-turn cold-start trajectory for the VSP task. }
    \label{apdxfig:data_sample_vsp}
    \vspace{-15pt}
\end{figure}

\begin{figure}[t!]
    \centering
    \includegraphics[width=1\textwidth]{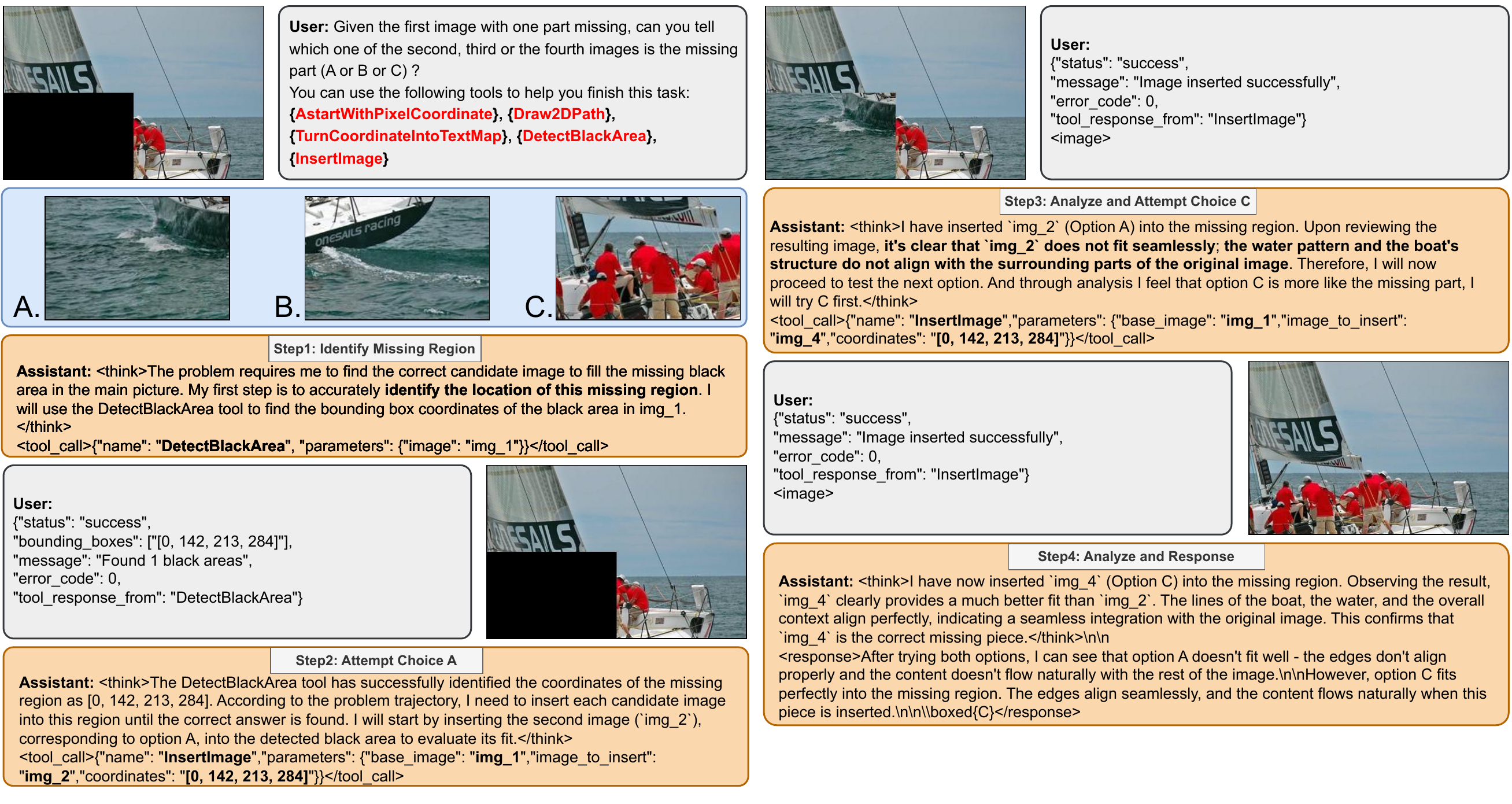}
    \caption{An example of a multi-turn cold-start trajectory for the Jigsaw task. This trajectory showcases an iterative trial-and-error process. The agent first uses \textsc{DetectBlackArea} to identify the missing region. It then sequentially attempts to InsertImage with each candidate patch, analyzing the visual result of each attempt before arriving at the correct solution.}
    \label{apdxfig:data_sample_jigsaw}
    % \vspace{-15pt}
\end{figure}

\begin{figure}[t!]
    \centering
    \includegraphics[width=1\textwidth]{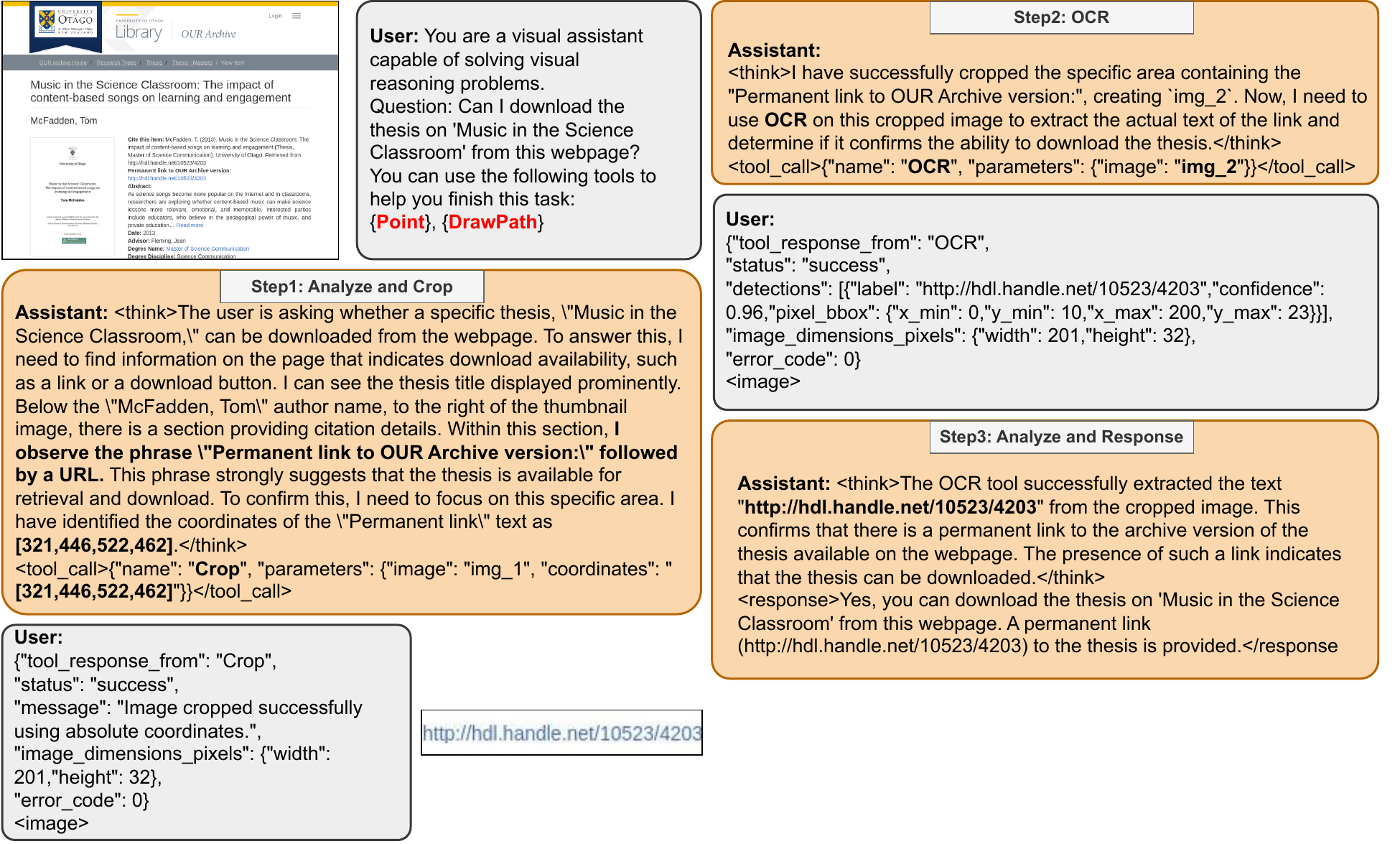}
    \caption{An example of a multi-turn cold-start trajectory for a GUI-QA task. This sample illustrates a focus-then-extract strategy. The agent first uses the \textsc{Crop} tool to isolate a specific, relevant section of the webpage. It then applies the \textsc{OCR} tool to this cropped, unambiguous input to perform precise information extraction.}
    \label{apdxfig:data_sample_web}
    % \vspace{-15pt}
\end{figure}

\subsection{High-Quality Cold Start Trajectory Data Curation}
\label{apdx:data_curation}
For our structured reasoning tasks, we developed customized data generation and trajectory creation pipelines to ensure high quality and diversity. Some detailed data samples are shown in appendix \ref{apdx:data_sample}

\paragraph{VSP}
The VSP benchmark environments were procedurally generated using the \textbf{Gymnasium}~\citep{towers2024gymnasium} framework. To ensure variety, we systematically controlled the distribution of start points, end points, and ice holes. We synthesized environments of sizes $4\times4$, $6\times6$, and $8\times8$ for the training set, while reserving larger $5\times5$, $7\times7$, and $9\times9$ grids for testing. The Tool Cold Start (TC, SFT) trajectories were designed to mimic an optimal problem-solving process. For Navigation tasks, the trajectory consists of: (1) invoking the \textsc{Point} tool to localize the start, end, and all ice holes; (2) performing textual reasoning based on these coordinates; and (3) calling \textsc{Draw2DPath} for final verification. Crucially, we also incorporated reflection and backtracking data derived from failure cases. For Verification tasks, the trajectory involves: (1) using \textsc{Point} to locate the start, (2) employing \textsc{Draw2DPath} to render the proposed path, and (3) prompting the model to judge if this path intersects any ice holes.

\paragraph{Jigsaw}
The Jigsaw dataset was constructed using images from the \textbf{COCO-2017}~\citep{cocopaper} training set. Each instance was created by first dividing an image into a $3\times3$ grid. A $2\times2$ sub-grid was then selected as the base image, from which one patch (e.g., top-right) was removed to create the problem. The removed patch served as the correct answer, while one of the remaining five patches from the original $3\times3$ grid was chosen as a distractor. The TC trajectory instructs the model to: (1) call \textsc{DetectBlackArea} to identify the coordinates of the missing section, and (2) iteratively call \textsc{InsertImage} for each candidate patch until the puzzle is solved. To enhance robustness and diversity, we introduced several key variations: \ding{202} The order of patch insertion attempts was randomized to ensure a uniform distribution of options. \ding{203} Scenarios involving tool failures (e.g., detection errors) were included, prompting the model to fall back on its intrinsic knowledge after several failed attempts. \ding{204} A proportion of samples were designed to be solvable directly by the model without tool use, encouraging adaptive tool invocation.

\paragraph{GUIQA}

The process begins with 44k single-turn instances from the \textbf{Guichat} dataset.~\citep{chen2024guicourse} To identify challenging cases that necessitate tool use, we first prompted a powerful vision-language model, Qwen-VL-2.5-72B~\citep{qwen25vl}, to answer the questions. We retained only the instances where the model failed, resulting in a subset of 7,100 ``hard'' questions. Next, for these 7,100 instances, we rendered the ground-truth answer coordinates as bounding boxes on the images. We then performed a manual visual inspection to ensure these boxes contained meaningful and relevant information, which filtered the set down to 1,800 valid data points. To generate high-fidelity tool-use trajectories for these cases, we provided the ground-truth answer and coordinates to gemini-2.5-flash~\cite{gemini25}, prompting it to produce the chain-of-thought reasoning and tool invocation sequence required to solve the problem. Finally, all generated trajectories were validated against our predefined format, and only those that strictly conformed were retained. This final curation step yielded a high-quality dataset of 1,139 instances for our TC stage.

After defining the abstract trajectory structure for all tasks, we followed a unified, two-stage process to create the final training data. First, we executed these trajectories programmatically to populate them with real tool inputs and outputs. Subsequently, we leveraged \textbf{Gemini 2.5 Flash}~\citep{gemini25} to generate the corresponding chain-of-thought (CoT) reasoning for each step. This process resulted in a final dataset of high-fidelity, tool-augmented trajectories complete with explicit reasoning chains, ready for our cold-start training.

\subsection{Data Samples}
\label{apdx:data_sample}
To provide a more concrete understanding of our cold-start data, we present representative multi-turn trajectory samples for each of our core tasks in Figures~\ref{apdxfig:data_sample_vsp}, \ref{apdxfig:data_sample_jigsaw}, and \ref{apdxfig:data_sample_web}. These examples are designed to showcase the sophisticated, human-like reasoning patterns we aim to instill in the model during the supervised fine-tuning phase.

The VSP sample (Figure~\ref{apdxfig:data_sample_vsp}) illustrates a methodical, multi-stage problem-solving process that includes perception, verification, and analysis. The Jigsaw sample (Figure~\ref{apdxfig:data_sample_jigsaw}) demonstrates an iterative trial-and-error strategy, where the agent actively evaluates the outcome of each tool call. Finally, the GUIQA sample (Figure~\ref{apdxfig:data_sample_web}) highlights a synergistic tool-use pattern, where one tool (`Crop') is used to create optimal conditions for another (`OCR'). Across all examples, the interplay between the model's internal thoughts (`$<$think$>$'), tool calls, and observations from the environment is clearly demonstrated.

\subsection{Tool GRPO}
\label{apdx:rl}
Group Relative Policy Optimization (GRPO) is a reinforcement learning algorithm that evaluates policy performance by directly comparing a group of candidate reasoning trajectories. The process of Tool GRPO in \method begins with the initial state $s_0 = \langle g, \text{text}, \text{image} \rangle$, for which the policy $\pi_\theta$ samples a set of $N$ complete trajectories as candidate responses, $\{\tau^1, \tau^2, \dots, \tau^N\}$. Each trajectory is evaluated by a reward function, yielding rewards $r^i = R(\tau^i)$. GRPO then calculates a group-relative advantage $A^i$ for each trajectory by normalizing its reward against the statistics of the entire group:
\begin{equation}
A^i = \frac{r_i - \operatorname{mean}\{r_1, r_2, \dots, r_N\}}{\operatorname{std}\{r_1, r_2, \dots, r_N\}}\,.
\end{equation}
The policy is then updated to favor trajectories with higher relative advantages by maximizing a clipped surrogate objective function. This objective is designed to ensure stable training by preventing excessively large policy updates. The full objective is:
\begin{equation}
\begin{aligned}
J_{\mathrm{GRPO}}(\theta)
&= \mathbb{E}_{q \sim P(Q),\{\tau^i\}_{i=1}^G \sim \pi_{\theta_{\mathrm{old}}}(\cdot|q)} \\
&\Biggl[
    \sum_{i=1}^G
    \sum_{j=1}^{|\tau^i|}
  \frac{1}{G|\tau^i|} 
  \min\!\Bigl(m^i_j A_i,\,
    \mathrm{clip}\!\Bigl(
      s^i_j,
      1-\varepsilon,\,
      1+\varepsilon
    \Bigr)
    A_i
  \Bigr)  -\,\beta\,\mathbb{D}_{\mathrm{KL}}\bigl(\pi_{\theta}\,\big\|\,\pi_{\mathrm{ref}}\bigr)
\Biggr]\,.
\end{aligned}
\label{eq1}
\end{equation}
Here, $m^i_j = \frac{\pi_\theta(\tau^i_j-s_i|s_i)}{\pi_{\theta_{old}}(\tau^i_j-s_i|s_i)}$ is the importance sampling ratio that measures the change between the new policy $\pi_\theta$ and the old policy $\pi_{\theta_{old}}$ used to generate the samples. The Kullback-Leibler (KL) divergence penalty, $\mathbb{D}_{KL}(\pi_\theta || \pi_{\text{ref}})$ regularizes the policy update by penalizing large deviations from a reference policy $\pi_{\text{ref}}$.

\paragraph{Reward Design}
Our reward function is designed to evaluate both the structural syntax and the semantic correctness of the model's output. The total reward, $R_{\text{total}}$, is a composite score defined as:
\begin{equation}
    R_{\text{total}} = R_{\text{format}} \cdot  (\lambda_{\text{tool}} \cdot  R_{\text{tool}} + \lambda_{\text{acc}} \cdot  R_{\text{acc}})
\end{equation}
Here, $R_{\text{format}}$ acts as a binary gate, ensuring that rewards for tool usage ($R_{\text{tool}}$) and final answer accuracy ($R_{\text{accuracy}}$) are granted only if the output adheres to the required structure. This design incentivizes the model to first master the correct syntax before optimizing for functional correctness. $\lambda_{\text{acc}}$ is a hyperparameter and the effect of it is discussed in Appendix~\ref{apdx:ablation_study}.

\begin{itemize}[leftmargin=*]
    \item \textbf{Format Reward ($R_{\text{format}}$)}
    The format reward is a binary signal that assesses the structural integrity of the model's output. It verifies that the generated response contains all required special tokens in the correct order and follows predefined rules.
    \begin{equation}
        R_{\text{format}} =
        \begin{cases}
            1 & \text{if the output format is valid} \\
            0 & \text{otherwise}
        \end{cases}
    \end{equation}
    If $R_{\text{format}}$ is 0, the total reward for the trajectory is nullified, creating a strong imperative for the model to learn the required output structure.
    
    \noindent\item \textbf{Tool Reward ($R_{\text{tool}}$)}
        The tool reward provides a fine-grained evaluation of the tool-calling process, with a score ranging from 0 to 4. We employ a hierarchical scoring system where each level must be passed to proceed to the next.
        
        \begin{enumerate}
            \item \textbf{Invocation Structure (Score 1):} A score of 1 is awarded if the tool call is correctly encapsulated within the \texttt{<tool\_call>} and \texttt{</tool\_call>} tokens. If not, the score is 0, and no further tool evaluation occurs.
            
            \item \textbf{Tool Name Validity (Score 2):} If the structure is correct, we verify that the invoked tool name exists in the set of available tools, $\mathcal{T}$. If the name is valid, the score becomes 2.
            
            \item \textbf{Parameter Name Correctness (Score $\in [2, 3]$):} If the tool name is valid, we assess the parameter names. A partial score is awarded based on the proportion of correctly named parameters. A perfect match yields a score of 3. The score is calculated as:
            \begin{equation}
                R_{\text{tool}} = 2 + \frac{|\text{params}_{\text{correct\_name}}|}{|\text{params}_{\text{total}}|}
            \end{equation}
            
            \item \textbf{Parameter Content Validity (Score $\in [3, 4]$):} Finally, if all parameter names are correct (base score of 3), we evaluate the parameter values for semantic and contextual validity. The final score is proportional to the number of correct values, reaching a maximum of 4.
            \begin{equation}
                R_{\text{tool}} = 3 + \frac{|\text{params}_{\text{correct\_content}}|}{|\text{params}_{\text{total}}|}
            \end{equation}
        \end{enumerate}
    
    \noindent\item \textbf{Accuracy Reward ($R_{\text{accuracy}}$)}
        The accuracy reward evaluates the final outcome of the reasoning process, providing a clear signal based on the correctness of the model's final answer.
        \begin{equation}
            R_{\text{accuracy}} =
            \begin{cases}
                4 & \text{if the final answer is correct} \\
                0 & \text{otherwise}
            \end{cases}
        \end{equation}
\end{itemize}
This multi-faceted reward structure guides the model toward not only achieving the correct final answer but also mastering the intermediate steps of correct formatting and precise tool invocation.

\section{Experiment Details}
\label{apdx:experiment_details}
\subsection{Task Definition}
\label{apdx:task_definition}
We evaluate our approach across a diverse suite of three challenging tasks to validate whether our approach can help enhance the reasoning ability.

\paragraph{Visual Spatial Planning} We adopt the FrozenLake scenario \citep{towers2024gymnasium} to evaluate models’ spatial planning and verification abilities. The \textbf{navigation} task requires the model to generate a safe path from the start to the goal while avoiding holes, which demands accurate perception of the grid map and robust sequential reasoning to plan multi-step trajectories. The \textbf{verification} task instead focuses on state checking, determining whether a given location or a proposed path is safe, which isolates the perception and reasoning components of the planning pipeline. Together, these tasks expose two fundamental challenges for VLMs: (i) precise visual perception of spatial layouts under varying map sizes, and (ii) reliable reasoning over action sequences to ensure safety and goal completion.

Concretely, we evaluate models on two benchmarks. The first is \textbf{VSPO}, a dataset we construct to assess out-of-distribution generalization. During training, we use maps of sizes $4\times4$, $6\times6$, and $8\times8$, while reserving maps of sizes $3\times3$, $5\times5$, $7\times7$, and $9\times9$ for testing. This setup not only probes the model’s ability to generalize to unseen spatial configurations but also examines whether it truly learns to leverage tool usage for problem solving. The second is the original \textbf{VSP benchmark} \citep{wu2024vsppaper}, where we adopt the navigation and verification tasks to further test visual-spatial reasoning and state-checking capabilities under standardized settings.
    
\paragraph{Jigsaw} The Jigsaw task \citep{jigsaworigin} evaluates a model’s ability to reconstruct holistic understanding from fragmented visual inputs. Specifically, the model must infer the correct spatial arrangement of shuffled image patches and reason about their part–whole relationships. This requires fine-grained perception to capture local details, as well as global reasoning to integrate them into a coherent whole. The key challenges lie in bridging local–global consistency and maintaining semantic alignment across patches, making it a strong test of visual compositionality and structural reasoning.

Concretely, we evaluate models on two benchmarks. The first is \textbf{Jigsaw-COCO}, where we construct training and test splits based on the COCO train 2017 dataset \citep{cocopaper}. We extract the top-left, top-right, and bottom-left patches of each image to form the training set, while reserving the bottom-right patches for testing. This design allows us to probe the model’s out-of-distribution generalization and examine whether it truly learns to invoke tool usage for solving the puzzle. The second is the \textbf{Jigsaw benchmark from BLINK (BLINK-J)} \citep{fu2024blink}, which provides a standardized evaluation of fine-grained visual reasoning and compositional understanding under more challenging and diverse settings.
    
\paragraph{GUIQA} The {GUIQA} task is designed to evaluate a model's sophisticated capabilities in fine-grained visual understanding and critical information extraction from GUIs. In this task, a model is provided with a GUI screenshot and an associated question, where the main difficulty lies in precisely grounding UI elements on a dense layout, comprehending their functional roles, and performing multi-step reasoning by integrating scattered information.
    
The evaluation is conducted on two distinct datasets. The first is the \textbf{GUIChat} \citep{chen2024guicourse}, which specifically probes the model's capacity for interactive, dialogue-based comprehension of webpage screenshots. Models are required to process complex queries related to visual information, human-centric needs, world knowledge, and reasoning. The second is the \textbf{WebQA} from the \textbf{WebMMU} \citep{awal2025webmmu}. It offers a structured evaluation across three distinct categories. Agentic Action tests the ability to understand UI elements like buttons and menus in order to formulate the necessary user actions, complete with precise spatial grounding. General Visual Comprehension assesses how well the model can extract and synthesize information from varied page components, including text, images, and graphics. And Multi-step Reasoning demands complex inference, numerical calculations, and comparisons across different parts of the UI. To better evaluate the effectiveness of our visual tools in GUI-based scenarios, we focus on the agent acting subset of the English split of WebMMU and report results on this subset in the main table to highlight the impact of our approach. Some of the complete WebMMU results are provided in Appendix~\ref{apdx:detailed_res}.

\paragraph{General VQA}
General VQA aims to evaluate the general visual ability of models beyond structured visual reasoning tasks, focusing on open-ended visual understanding and reasoning in unconstrained real-world scenarios. Unlike structured benchmarks that provide well-defined goals and tool usage patterns, general VQA tasks often require flexible perception, commonsense reasoning, and robust interpretation of diverse visual content without explicit procedural guidance.

To assess the general visual capabilities of our approach in such settings, we consider two representative benchmarks: \textbf{V*}\citep{wu2024vstar} and \textbf{HRBench}\citep{hrbench}. \textbf{V*} evaluates a model’s ability to perform fine-grained visual search and object attribute recognition in high-resolution, complex visual scenes, focusing on precise localization and relationship reasoning that goes beyond simple image understanding. \textbf{HRBench} assesses the perception and reasoning capabilities of Multimodal Large Language Models on ultra-high-resolution images (up to 8K), testing fine-grained single-instance and cross-instance visual understanding, including detailed attribute extraction and spatial relationships in larger scenes.

\subsection{Prompts}
\label{apdx:system_prompt}
The system prompt used for guiding the tool-planning model is provided in Figure \ref{apdxfig:system_prompt}.

\begin{figure}[htbp]
    \centering
\begin{promptbox}{System Tool Prompt}

You are a visual assistant capable of solving visual reasoning problems. You can rely on your own capabilities or use external tools to assist in solving. \\
Available Tools  
In your response, you can use the following tools:  

\{\textcolor{red}{Tool List}\}\\
Steps for Each Turn\\
1. \textbf{Think:} First, silently analyze the user's request to understand the goal. This thinking process should be enclosed in \tooltoken{<think>} and \tooltoken{</think>} tags.\\
2. \textbf{Decide Action:} Based on your thinking, decide on one of the following two actions:\\
   - \textbf{If you need to use a tool:} Generate your tool call, enclosed between \tooltoken{<tool\_call>} and \tooltoken{</tool\_call>} tags. \textbf{Do not} generate a \tooltoken{<response>} in this turn.\\
   - \textbf{If you have enough information to answer:} Generate your final, user-facing answer, enclosed between \tooltoken{<response>} and \tooltoken{</response>} tags. \textbf{Do not} generate a \tooltoken{<tool\_call>} in this turn.\\
Output Format:\\
Your output must always begin with your thought process. After the \tooltoken{<think>} block, you must provide either a \tooltoken{<tool\_call>} or a \tooltoken{<response>}, but \textbf{never both} in the same turn.\\
\textbf{Case 1: Tool Use is Required}\\
\tooltoken{<think>} Your thoughts and reasoning \tooltoken{</think>} \\ 
\tooltoken{<tool\_call>}  \\
\{``name'': ``Tool name'', ``parameters'': \{``Parameter name'': ``Parameter content'', ``...'': ``...''\}\}\\
\tooltoken{</tool\_call>}  \\
\textbf{Case 2: Ready to Respond to the User}\\
\tooltoken{<think>} Your thoughts and reasoning \tooltoken{</think>}\\  
\tooltoken{<response>} Your final response \tooltoken{</response>}\\
Important Notes  \\
1. You must always include the \tooltoken{<think>} field to outline your reasoning. Provide one of \tooltoken{<tool\_call>} or \tooltoken{<response>}. You must not include both \tooltoken{<tool\_call>} and \tooltoken{<response>} in the same turn because they are mutually exclusive. If tool usage is required, you must instead include both \tooltoken{<think>} and \tooltoken{<tool\_call>}, and omit \tooltoken{<response>} for that turn. If no further tool usage is required and ready to answer the user's question, you can then use \tooltoken{<think>} to summarize your reasoning and include \tooltoken{<response>} with your final answer, and this indicates the ends of the conversation.\\
2. You can only invoke a single tool call at a time in the \tooltoken{<tool\_call>} fields. The tool call should be a JSON object with a ``name'' field and a ``parameters'' field containing a dictionary of parameters. If no parameters are needed, leave the "parameters" field an empty dictionary. All images have their coordinate origin at the top-left corner.\\
3. Some tools require image input. You do not need to generate or upload the actual image data simply refer to an image using a placeholder in the form of ``img\_n''. There may be multiple images present in the dialogue. Besides the original image, additional images may appear as a result of prior tool calls (e.g., edited images returned by visual editing tools). You are free to select which image to use as input for the next tool.
The index n in ``img\_n'' refers to the image's position in the dialogue history:\\
- The original image is always referred to as ``img\_1''.\\
- Each subsequent image, including those returned from tools, is assigned ``img\_2'', ``img\_3'', and so on, in the order they appear in the dialogue.\\
For example:\{``parameters'': \{``image'': ``img\_1'', ``other\_params'': ``other\_values''\}\}\\
4. All image coordinates used must be in absolute pixel values, not relative or normalized coordinates. \\
5. At the end, provide your final answer by placing it inside \\boxed\{\}, and wrap the entire final output inside \tooltoken{<response>}\tooltoken{</response>} tags.\\

\end{promptbox}
    \caption{Our system employs tool prompts to guide models in learning how to use tools effectively.}
    \label{apdxfig:system_prompt}
\end{figure}

\subsection{Implementation Details}
\label{apdx:implement_details}
We developed \textbf{AdaReasoner Framework}, an end-to-end framework that orchestrates the entire lifecycle of our tool-planning models, from data curation to evaluation. At the heart of this framework is the Tool Server, a unified, MCP-like service that manages all available tools, from simple offline utilities to compute-heavy online expert models.

\subsubsection{Data Curation}
During the data curation stage, we employ our \textbf{AdaDataCuration} module, which leverages the Tool Server to automatically generate high-quality cold-start trajectories. Specifically, we first design abstract, optimal problem-solving blueprints for each task, consisting of a tool-call chain and chain-of-thought (CoT) placeholders. We then prompt Gemini-2.5-Flash to fill these placeholders with detailed CoT reasoning. Finally, the Tool Server executes the corresponding tool calls and integrates the results into the dialogue, yielding a complete and coherent training instance.

\subsubsection{Tool Cold Start Stage}

During the cold-start stage, these trajectories are used for full-parameter supervised fine-tuning, for which we adopt the LLaMA Factory framework \citep{zheng2024llamafactory}. The key configurations and hyperparameters are summarized in Table \ref{tab:coldstart-config}.

\begin{table}[t]

\centering
\small
\begin{tabular}{llc}
\toprule
\textbf{Category} & \textbf{Hyperparameter} & \textbf{Value / Setting} \\
\midrule
\multirow{4}{*}{\textbf{Model}} 
& Base Model & Qwen2.5-VL-7B-Instruct \\
& Vision Tower Frozen & True \\
& MM Projector Frozen & True \\
& Finetuning Type & Full \\
& DeepSpeed Stage & ZeRO-3 \\
\midrule
\multirow{3}{*}{\textbf{Dataset}}
& Max Samples & 332,649 \\
& Cutoff Length & 35,536 \\
& Preprocessing Workers & 64 \\
\midrule
\multirow{7}{*}{\textbf{Training}}
& Batch Size per Device & 1 \\
& Gradient Accumulation Steps & 2 \\
& Effective Batch Size & 2 \\
& Learning Rate & 1e-5 \\
& Epochs & 3 \\
& LR Scheduler & cosine \\
& Warmup Ratio & 0.1 \\
& Mixed Precision & bfloat16 \\
\midrule
\multirow{2}{*}{\textbf{Logging / IO}}
& Logging Steps & 10 \\
& Checkpoint Save Steps & 100 \\
\midrule
\multirow{3}{*}{\textbf{Evaluation}}
& Train/Validation Split & 90\% / 10\% \\
& Eval Batch Size per Device & 1 \\
& Eval Steps & 100 \\
\bottomrule
\end{tabular}
\caption{Tool Cold Start (SFT) Training Configuration and Hyperparameters.}
\label{tab:coldstart-config}
\end{table}

\subsubsection{Tool GRPO Stage}
Following SFT, the model is further refined in the Tool-GRPO stage using \textbf{AdaTG}, our custom reinforcement learning framework inspired by \cite{verl_bytedance, deepeyes}, which also relies on the Tool Server for live tool interactions.
Specifically, AdaTG enables online reinforcement learning by allowing the model to interact with tools in real time through the Tool Server, receiving execution feedback at each planning step. Combined with the Tool-GRPO optimization objective, this interactive training process explicitly aligns the policy with long-horizon tool planning behaviors, resulting in a robust and adaptive tool planning model. The key configurations and hyperparameters of Tool GRPO stage are summarized in Table \ref{tab:rl-hparams}.

\begin{table}[t]

\centering
\small
\begin{tabular}{lll}
\toprule
\textbf{Category} & \textbf{Hyperparameter} & \textbf{Value / Setting} \\
\midrule

\multirow{5}{*}{\textbf{Data}} 
 & Max Prompt Length & 8192 tokens \\
 & Max Response Length & 20480 tokens \\
 & Train Batch Size & 32 \\
 & Shuffle & True \\
 & Filter Overlong Prompts & True \\

\midrule
\multirow{13}{*}{\textbf{Policy}} 
 & Strategy & FSDP \\
 & Gradient Checkpointing & True \\
 & PPO Mini-batch Size & 8 \\
 & PPO Micro-batch Size / GPU & 1 \\
 & Max Token Len / GPU (PPO) & 16384 \\
 & Grad Clip & 1.0 \\
 & Clip Ratio (PPO) & 0.2 \\
 & PPO Epochs & 1 \\
 & Entropy Coeff & 0.0 \\
 & Use KL Loss & False \\
 & Actor LR & 1e-6 \\
 & Weight Decay & 0.01 \\
 & FSDP Param Offload & True \\
 & FSDP Optimizer Offload & True \\
  & \# Nodes / GPUs & 1 node, 8 GPUs \\

\midrule
\multirow{11}{*}{\textbf{Rollout}} 
 & Engine & vLLM \\
 & Temperature & 1.0 \\
 & Top-p & 1.0 \\
 & Top-k & -1 \\
 & \# Samples per Prompt (n) & 4 \\
 & Dtype & bfloat16 \\
 & Tensor Model Parallel Size & 2 \\
 & Max \# Batched Tokens & 32768 \\
 & GPU Memory Utilization & 0.65 \\
 & Enforce Eager & False \\
 & Chunked Prefill & False \\

\midrule
\multirow{1}{*}{\textbf{Tool-Agent}}
 & Max Turns per Episode & 10 \\

\midrule
\multirow{6}{*}{\textbf{Critic}} 
 & Strategy & FSDP \\
 & LR & 1e-5 \\
 & Weight Decay & 0.01 \\
 & PPO Epochs & 1 \\
 & Grad Clip & 1.0 \\

\midrule
\multirow{6}{*}{\textbf{Algorithm}} 
 & Advantage Estimator & GRPO \\
 & Gamma & 1.0 \\
 & Lambda & 1.0 \\
 & Use KL in Reward & False \\
 & KL Coef & 0.0 \\
 & Norm Adv by Std in GRPO & True \\

\bottomrule
\end{tabular}
\caption{Key configurations and hyperparameters used in the Tool GRPO stage.}
\label{tab:rl-hparams}
\end{table}

\subsubsection{Details of Our Final Model}
\label{apdx:final_model_details}
During the training of our final model, \textbf{AdaReasoner-7B}, we adopt a two-stage training pipeline consisting of a cold-start stage followed by a Tool-GRPO stage. In the \textbf{Tool Cold Start (TC) stage}, we fine-tune the model using data from \textbf{VSP}, \textbf{Jigsaw}, and \textbf{WebQA}. During this phase, we enable \textbf{Adaptive Learning} to facilitate efficient adaptation across diverse task structures and reasoning patterns.
In the subsequent \textbf{Tool GRPO (TG) stage}, we further train the model using a broader set of tasks, including \textbf{VSP}, \textbf{Jigsaw}, \textbf{WebQA}, and additional \textbf{visual search} data. Adaptive Learning remains enabled throughout this stage to continuously adjust the model’s behavior as it interacts with different tools and task distributions.
This two-stage training strategy allows AdaReasoner-7B to progressively acquire strong reasoning capabilities while maintaining robustness and generalization across diverse multimodal tasks.

In Table \ref{tab:tool_statistics_multi_task}, we report the performance differences of several prior methods under their original evaluation settings and our unified new setting. We observe that the performance gaps mainly stem from changes in tool types and definitions, as well as differences in the input order of images and questions. In contrast, our model exhibits strong robustness on the V* benchmark: its performance remains stable even when the image–question order is swapped or when the tool definition is changed, demonstrating superior adaptability to variations in tool definitions and evaluation protocols.

\subsection{Evaluation Details}
\subsubsection{Baselines}
We evaluate our method against a comprehensive set of strong baselines.
First, we include proprietary models with state-of-the-art multimodal capabilities, including GPT-5-20250807~\citep{openai2025gpt5}, Claude-sonnet-4-20250514~\citep{anthropic2025claude4sonnet}, and Gemini-2.5-flash~\citep{gemini25}.
Second, we consider open-source MLLMs, namely Qwen-2.5-VL-32B/72B-Instruct~\citep{qwen25vl} and InternVL-3-78B~\citep{zhu2025internvl3}, which are selected as primary testbeds due to their strong performance in visual understanding and reasoning, enabling us to assess the scalability of our approach across different model sizes.

\subsubsection{Evaluation Method}
We design \textbf{AdaEval}, a unified evaluation framework for tool planning that supports the evaluation of both tool-planning models and non–tool-planning models.
To ensure fair and consistent comparison, all models in our experiments are evaluated exclusively within the AdaEval framework under the same evaluation protocol.

For benchmarks that require open-ended or subjective judgment, including VStar, WebMMU, and GUIQA, we adopt an LM-as-a-Judge evaluation strategy. Specifically, we use Qwen-2.5-VL-72B as the judge model. The prompt used for judgment is illustrated in Figure~\ref{apdxfig:lm_as_ajudge}.

\begin{figure}[htbp]
    \centering
\begin{promptbox}{System Tool Prompt}

You are an expert evaluator. Your goal is to determine if a [Model Answer] correctly and factually answers a [Question] when compared against a [Standard Answer].

\textbf{Core Evaluation Principle:}

The [Model Answer] is considered consistent if it contains the \textbf{essential key information} present in the [Standard Answer]. The [Model Answer] is allowed to be much more verbose, conversational, and include additional correct context or explanations. Your primary task is to \textbf{verify the presence of the core facts}, not to penalize extra information. If a question asks for specific formatting like coordinates or tables, but the model identifies the correct core element textually, it should still be considered consistent.

- \textbf{Consistent (Judgement: 1):} The [Model Answer] successfully identifies the main point or action from the [Standard Answer]. For example, if the standard answer is to ``click button A'', the model answer is consistent if it mentions clicking or interacting with ``button A'', even if it's surrounded by other text.

- \textbf{Inconsistent (Judgement: 0):} The [Model Answer] fails to mention the key information, provides contradictory information, or hallucinates a different solution.

\textbf{Output Format:}

Just output `Judgement: 1' or `Judgement: 0'. Do not output anything else.

\end{promptbox}
    \caption{System Prompt Used for the LM-as-a-Judge Evaluation}
    \label{apdxfig:lm_as_ajudge}
\end{figure}

In addition, we conduct a human evaluation on the V* benchmark by manually assessing the extracted model predictions of Qwen 2.5 VL 7B and AdaReasoner-7B. The human judgment scores are consistent with the scores produced by our evaluation framework, providing strong evidence for the reliability and correctness of our automated evaluation.

\subsubsection{Detailed Results}
\label{apdx:detailed_res}
Due to space constraints and for clarity of presentation, we report abridged results in the main paper and provide the full results here.
Specifically, in Section~\ref{sec:exp_single_task_train}, we evaluate the contributions of TC and TG under single-task fine-tuning. The summarized results are reported in Table~\ref{tab:main_table}, with the corresponding detailed results provided in Table~\ref{tab:part4_full}.
In Section~\ref{sec:generalization_1_to_3}, we present the generalization performance in Table~\ref{tab:generalization_1_to_3}, and report the full benchmark breakdown in Table~\ref{tab:generalization_full}.
Finally, in Section~\ref{sec:exp_main_res}, we report the main results in Table~\ref{tab:part4_main_table}, with the complete version shown in Table~\ref{tab:part4_full}.
\subsection{Ablation Study}
\label{apdx:ablation_study}

We systematically adjust $\lambda_{\text{tool}}$ and $\lambda_{\text{acc}}$ to evaluate their influence on learning dynamics and final performance. Specifically, we train the model on the same VSP task data for 100 RL steps under different reward-weight settings, monitor the training curves to ensure convergence, and then evaluate each checkpoint's performance. The results are summarized in the table \ref{tab:reward_ablation}.

\begin{wrapfigure}{r}{0.5\textwidth}
\vspace{-10pt}
\centering

\resizebox{0.48\textwidth}{!}{
\begin{tabular}{c|ccc|ccc}
\toprule
\multicolumn{1}{c|}{$\lambda_{\text{tool}}:\lambda_{\text{acc}}$} &
\multicolumn{3}{c|}{VSP (\%)} &
\multicolumn{3}{c}{VSPO (\%)} \\
\cmidrule(lr){2-4} \cmidrule(lr){5-7}
& Nav & Verify & Overall & Nav & Verify & Overall \\
\midrule
\textbf{0:1} & 51.83 & 95.00 & 71.45 & 41.78 & 75.58 & 57.37 \\
\textbf{1:2} & 49.50 & 95.80 & 70.55 & 36.44 & 94.29 & 63.11 \\
\textbf{1:1} & 64.00 & 96.40 & 78.73 & 48.56 & 96.23 & 70.54 \\
\textbf{2:1} & \textbf{90.33} & \textbf{96.80} & \textbf{93.27} &
      \textbf{70.33} & \textbf{96.36} & \textbf{82.34} \\
\bottomrule
\end{tabular}
}
\captionof{table}{Ablation on reward-weight configurations for VSP and VSPO.}
\label{tab:reward_ablation}
\end{wrapfigure}

As shown in table \ref{tab:reward_ablation}, the model's performance consistently improves as the ratio $\lambda_{\text{tool}} : \lambda_{\text{acc}}$ increases. This indicates that larger tool rewards not only accelerate convergence during RL training but also lead to significantly better final performance. These results validate that our tool-reward design is effective and plays a crucial role in helping the model learn tool calling more efficiently and robustly.

\section{Further Analysis}

\begin{wrapfigure}{r}{0.4\textwidth}
\vspace{-15pt} 
\centering
% \setlength{\tabcolsep}{4pt} % 调整列间距
% \begin{figure}[t]
    \centering
    \includegraphics[width=\linewidth]{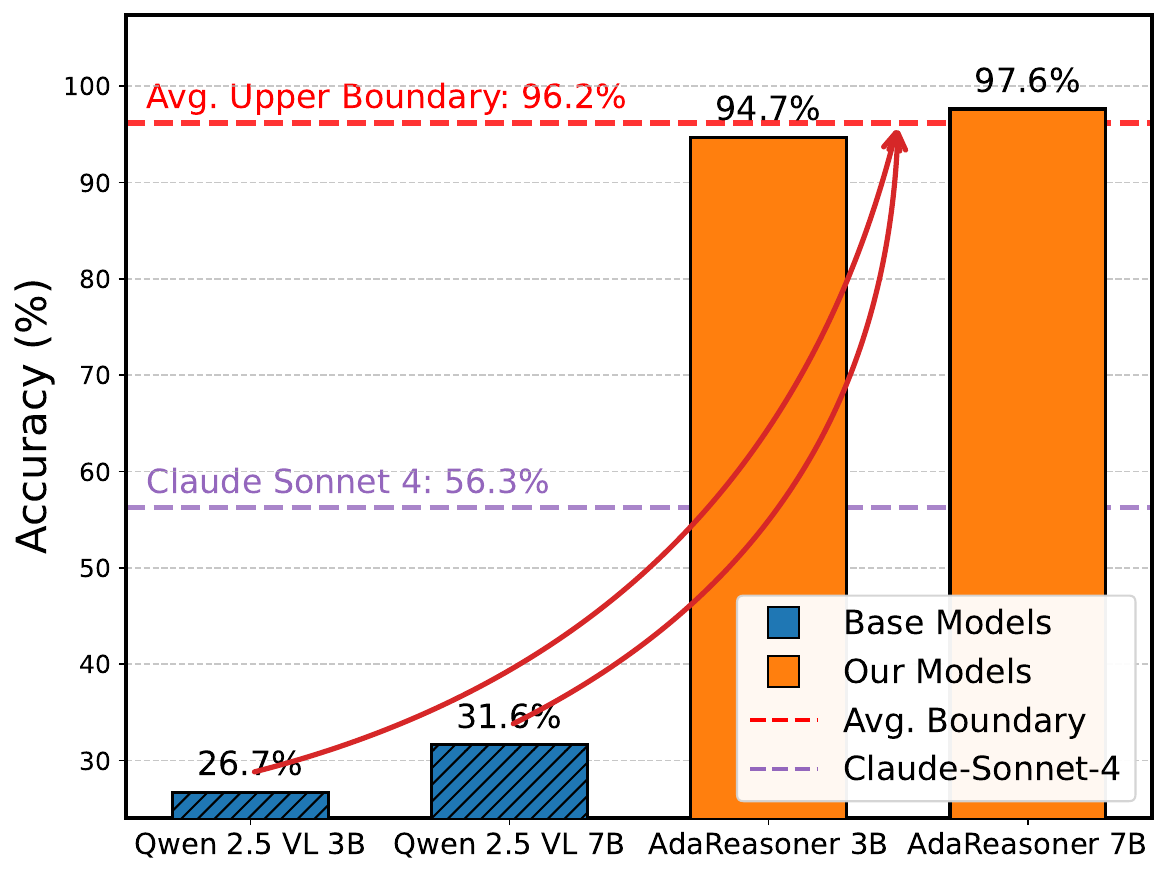}
    \caption{Overcoming scale-based limitations with tool augmentation. On the VSP task, our tools boost the performance of both 3B and 7B models, elevating them from disparate baselines to a near-uniform high performance.}
\label{fig:tool_model_boundary}
\vspace{-12pt} 
\end{wrapfigure}

\subsection{Visual Tools Help Overcome Scale-Based Limitations} 

The results shown in section \ref{sec:exp_single_task_train} reveal that tool augmentation can redefine the performance ceiling of MLLMs by overcoming scale-based limitations. As illustrated in Figure~\ref{fig:tool_model_boundary}, while the baseline performance of 3B and 7B models is disparate and low, our tool-augmented versions both achieve near-perfect accuracy (94.7\% and 97.6\%). This indicates that the primary performance bottleneck has shifted from the model’s intrinsic scale to the extrinsic quality of the tools it wields. Consequently, this establishes a powerful paradigm where even smaller, more efficient models can achieve state-of-the-art results, contingent not on their size, but on the instruments they are equipped with.

\subsection{Why Visual Tools Help}
\label{apdx:why_visual_tools_help}

Our framework decomposes complex reasoning tasks into manageable steps, each resolved either by the model itself or by a high-precision external tool. This design fundamentally \emph{shifts the problem-solving burden}: instead of requiring flawless internal reasoning, the model's primary task becomes effective tool planning. By delegating precise sub-tasks to reliable tools, the model is freed to focus on its core competencies of judgment, synthesis, and integrating the resulting outputs.

\begin{table*}[h]
\centering
\small
\tablestyle{14pt}{1.0}
\resizebox{.98\textwidth}{!}{%
\begin{tabular}{l|c|ccc}
\toprule
\multirow{2}{*}{\textbf{Model}} & \textbf{Perception} & \multicolumn{3}{c}{\textbf{VSP-Verify Reasoning Accuracy}} \\
\cmidrule(lr){2-2} \cmidrule(lr){3-5}
& Point Loc. Acc. $\uparrow$ & Base & w/ Line & w/ Point \\
\midrule
Qwen 2.5 VL 3B  & 2.47   & 50.91 & 57.92 ({\color{ForestGreen}+7.01}) & 49.09 ({\color{red}-1.82}) \\
Qwen 2.5 VL 7B  & 47.01  & 50.85 & 57.68 ({\color{ForestGreen}+6.83}) & 57.87 ({\color{ForestGreen}+7.02}) \\
Qwen 2.5 VL 32B & 6.54   & 53.12 & 61.31 ({\color{ForestGreen}+8.19}) & 87.87 ({\color{ForestGreen}+34.75}) \\
Qwen 2.5 VL 72B & 50.00  & 52.34 & 61.57 ({\color{ForestGreen}+9.23}) & 87.53 ({\color{ForestGreen}+35.19}) \\
\midrule
\rowcolor{gray!12}
\textbf{Our \textsc{Point} Tool} & \textbf{100.0} & -- & -- & \textbf{--} \\
\bottomrule
\end{tabular}
}
\caption{\textbf{From Perception to Reasoning: Impact of our \textsc{Point} tool.} 
(Left) Comparison of start-point localization accuracy between Qwen 2.5 VL models and our specialized tool. 
(Right) Zero-shot reasoning performance on VSP-Verify task when augmented with tool-generated visual context (Line/Point). 
The significant gains in 32B/72B models demonstrate that high-precision perception is the key to unlocking agentic reasoning.}
\label{tab:combined_perception_reasoning}

\end{table*}

\paragraph{Perception Tools Help MLLMs to See} Our framework leverages expert perception tools to overcome the intrinsic perceptual limitations of MLLMs. As shown in Table \ref{tab:combined_perception_reasoning}, in VSP-verification, our expert \textsc{Point} tool achieves perfect localization accuracy (100.0\% vs. $\sim50.0\%$ for baselines), and providing its coordinate output as context boosts the downstream zero-shot reasoning performance by an average of \textbf{\textcolor{ForestGreen}{+18.79}} points. This principle holds even with imperfect tools: for the Jigsaw task, our \textsc{DetectBlackArea} tool achieves only \textbf{72.6\%} accuracy, yet it still provides a significant perceptual advantage, underscoring the value of delegating such challenges to specialized tools.

\paragraph{Manipulating Tools help MLLMs to verify} Our manipulating tools empower the model to formulate and subsequently verify its own hypotheses. For example, in the VSP-Verify task, we teach the model to call \textsc{Draw2DPath} to explicitly draw a red line on the frozen lake question picture. The problem is thus converted to verifying whether the red line crosses the blue ice holes. As shown in Table \ref{tab:combined_perception_reasoning}, even under a zero-shot context-appending setting, the \textsc{DrawLine} does help improve the judge accuracy of the model, yielding an average performance improvement of \textbf{\textcolor{ForestGreen}{+7.82}} points. 
Similarly, in the Jigsaw task, the model can actively invoke the \textsc{InsertImage} tool to compose and evaluate different candidate pieces by inserting them into the original image, enabling more informed decision-making.

\paragraph{High-Quality Trajectory data help MLLMs to plan} While augmenting context with tool outputs is effective for zero-shot reasoning, this strategy alone is insufficient for achieving optimal performance. Tool-Cold-Start addresses this gap by explicitly teaching the model two foundational capabilities: \emph{how to use tools correctly} and \emph{how to recognize the patterns} where they should be applied. As shown in Table \ref{tab:main_table}, for the 7B models, adding the Tool-Cold-Start phase before Tool-GRPO yields a massive performance improvement of \textbf{\textcolor{ForestGreen}{+24.93}} points on VSP and \textbf{\textcolor{ForestGreen}{+19.82}} points on Jigsaw compared to using Tool-GRPO alone. Besides this, the inclusion of \emph{reflection} data during the Cold-Start phase provides further benefits to the model's reasoning. As shown in Table \ref{tab:adaptive_study}, when \textsc{AStar} search is disabled, training with reflection data yields a substantial improvement over the no-reflection checkpoints (91.36 vs. 67.27).

\subsection{The Dual Role of Cold-Start Supervision}

A central finding of our work concerns the dual role of the Tool Cold Start (TC, SFT) phase, which highlights a critical trade-off between imparting expert knowledge and preserving a model's exploratory freedom. Our results suggest that the decision to include a supervised pre-training stage is not universally beneficial, but rather highly contingent on the nature of the task.

For complex, structured tasks with discernible optimal solutions, such as VSP and Jigsaw, the SFT phase provides a decisive advantage. In these scenarios, discovering an effective tool-use trajectory from scratch is a non-trivial exploration problem for the model due to its inherent reasoning or knowledge deficits. By exposing the model to high-quality, deterministic solution paths, the Tool Cold Start phase effectively bootstraps the learning process, instilling a strong inductive bias towards a correct strategy. The empirical results in Table~\ref{tab:main_table} validate this unequivocally: for our 7B models, adding this SFT phase before Tool-GRPO yields a massive performance improvement of \textbf{\textcolor{ForestGreen}{+24.93}} points on VSP and \textbf{\textcolor{ForestGreen}{+19.82}} points on Jigsaw compared to using Tool-GRPO alone.

Conversely, for open-ended and highly generalized domains like GUIQA, the limitations of this pre-defined guidance become apparent. In such settings, the optimal tool-use strategy is often unknown even to human designers, making any human-designed trajectory likely sub-optimal. We find that a rigid SFT phase can inadvertently restrict the model's exploratory freedom during subsequent RL by creating a strong policy bias, which hinders the discovery of more effective, emergent strategies. This effect is clearly observed in our results for the 7B model on the WebMMU benchmark, where the standalone Tool-GRPO approach actually outperforms the combined pipeline (\textbf{72.97 vs. 68.16}).

This dichotomy suggests a key principle for training tool-augmented agents: while injecting expert knowledge via SFT is a powerful method for tasks with well-defined solution spaces, a pure reinforcement learning approach like Tool-GRPO may be superior for more dynamic and general tasks that benefit from unconstrained exploration.

\begin{table*}[t]
\centering
\small
\setlength{\tabcolsep}{5pt}
\renewcommand{\arraystretch}{1.15}
\resizebox{\textwidth}{!}{
\begin{tabular}{
l
| S[table-format=2.2] S[table-format=2.2] S[table-format=2.2]
| S[table-format=2.2] S[table-format=2.2] S[table-format=2.2]
| S[table-format=2.2] S[table-format=2.2] S[table-format=2.2] S[table-format=2.2]
}
\toprule
\multirow{2}{*}{\textbf{Model}} &
\multicolumn{3}{c|}{\textbf{VSPO}} &
\multicolumn{3}{c|}{\textbf{VSP}} &
\multicolumn{4}{c}{\textbf{WebMMU}} \\
\cmidrule(lr){2-4}\cmidrule(lr){5-7}\cmidrule(lr){8-11}
& {Nav} & {Verify} & {Overall}
& {Nav} & {Verify} & {Overall}
& {Avg.} & {Act.} & {Comp.} & {Reason.} \\
\midrule

Qwen2.5-VL 3B            & 5.67 & 50.91 & 26.53 & 7.50 & 49.80 & 26.73 & 45.39 & 55.89 & 51.82 & 34.95 \\
\;\;+ Direct SFT             & 27.42 & 49.66 & 38.15 & 34.50 & 44.00 & 38.82 & 46.54 & 61.38 & 54.46 & 32.31 \\
\;\;+ Direct GRPO            & 2.78 & 50.00 & 24.55 & 18.33 & 50.00 & 32.73 & 48.44 & 56.30 & 51.49 & 41.41 \\
\;\;+ TC (cold-start)        & 14.67 & 84.81 & 47.01 & 23.33 & 84.40 & 51.09 & 35.03 & 44.72 & 42.24 & 24.82 \\
\;\;+ TG (tool-GRPO)         & 11.22 & 50.00 & 29.10 & 22.67 & 50.00 & 35.09 & \underline{58.88} & \underline{72.15} & \underline{62.05} & 47.87 \\

\rowcolor{gray!12}
\;\;\textbf{+ TC + TG} &
\underline{73.00} & \underline{98.44} & \underline{84.73} &
\underline{92.17} & \underline{97.80} & \underline{94.73} &
\textbf{63.48} & \textbf{81.71} & \textbf{57.43} & \underline{53.01} \\

\rowcolor{gray!6}
$\Delta$ vs.\ base &
\color{ForestGreen}+67.33 & \color{ForestGreen}+47.53 & \color{ForestGreen}+58.20 &
\color{ForestGreen}+84.67 & \color{ForestGreen}+48.00 & \color{ForestGreen}+68.00 &
\color{ForestGreen}+18.09 & \color{ForestGreen}+25.82 & \color{ForestGreen}+5.61 & \color{ForestGreen}+18.06 \\

\midrule

Qwen2.5-VL 7B            
& 5.22 & 48.96 & 25.39 
& 12.33 & 47.00 & 28.09 
& 59.08 & 67.48 & \textbf{69.31} & 48.46 \\

\;\;+ Direct SFT             
& 33.68 & 51.30 & 42.18 
& 42.67 & 51.40 & 46.64 
& 55.62 & 65.65 & 63.70 & 44.79 \\

\;\;+ Direct GRPO            
& 10.33 & 49.48 & 28.38 
& 12.50 & 51.40 & 30.18 
& \underline{70.19} & \underline{83.54} & \underline{69.31} & \underline{60.94} \\

\;\;+ TC (cold-start)        
& 31.58 & \underline{94.01} & \underline{61.69} 
& 41.00 & \underline{93.60} & \underline{64.91} 
& 51.63 & 64.63 & 54.13 & 41.12 \\

\;\;+ TG (tool-GRPO)         
& \underline{65.89} & 52.47 & 59.70 
& \underline{88.17} & 55.20 & 73.18 
& \textbf{72.97} & \textbf{88.62} & 66.34 & \textbf{64.61} \\

\rowcolor{gray!12}
\;\;\textbf{+ TC + TG} &
\textbf{73.44} & \textbf{98.70} & \textbf{85.09} &
\textbf{96.33} & \textbf{99.20} & \textbf{97.64} &
68.16 & 82.32 & 67.33 & 58.30 \\

\rowcolor{gray!6}
$\Delta$ vs.\ base &
\color{ForestGreen}+68.22 &
\color{ForestGreen}+49.74 &
\color{ForestGreen}+59.70 &
\color{ForestGreen}+84.00 &
\color{ForestGreen}+52.20 &
\color{ForestGreen}+69.55 &
\color{ForestGreen}+9.08 &
\color{ForestGreen}+14.84 &
\color{red}-1.98 &
\color{ForestGreen}+9.84 \\

\bottomrule
\end{tabular}
}
\caption{\textbf{Sub-task Breakdown for Single Task Study.} 
Detailed results for VSPO (Navigation and Verification), VSP (Navigation and Verification), and WebMMU (Avg.\ = overall average, Act.\ = Agentic Action, Comp.\ = Visual Comprehension, Reason.\ = Multi-step Reasoning). 
Best is \textbf{bold}, second-best is \underline{underlined}. 
See Table~\ref{tab:main_table} for complete results including Jigsaw, BLINK-J, and GUIChat.}
\label{tab:full_results}
\end{table*}

\begin{table*}[t]
\centering
\small
\setlength{\tabcolsep}{5pt}
\renewcommand{\arraystretch}{1.05}

\resizebox{\textwidth}{!}{
\begin{tabular}{
l
| S[table-format=2.2] S[table-format=2.2] S[table-format=2.2]
| S[table-format=2.2] S[table-format=2.2] S[table-format=2.2]
| S[table-format=2.2] S[table-format=2.2] S[table-format=2.2] S[table-format=2.2]
}
\toprule
\multirow{2}{*}{\textbf{Model}} &
\multicolumn{3}{c|}{\textbf{VSPO}} &
\multicolumn{3}{c|}{\textbf{VSP}} &
\multicolumn{4}{c}{\textbf{WebMMU}} \\
\cmidrule(lr){2-4}\cmidrule(lr){5-7}\cmidrule(lr){8-11}
& {Nav} & {Verify} & {Overall}
& {Nav} & {Verify} & {Overall}
& {Avg.} & {Act.} & {Comp.} & {Reason.} \\
\midrule

Qwen2.5-VL 7B            & 5.22 & 48.96 & 25.39 & 12.33 & 47.00 & 28.09 & 59.08 & 67.48 & \textbf{69.31} & 48.46 \\
\;\;+ TC                     & 9.11 & 49.35 & 27.66 & 12.83 & 49.20 & 29.36 & 43.16 & 41.87 & 58.75 & 37.15 \\
\;\;+ Rnd TC                 & 7.33 & 49.35 & 26.71 & 14.17 & 49.80 & 30.36 & 40.51 & 42.89 & 52.15 & 33.63 \\
\;\;+ TG                     & 1.78 & 50.00 & 24.01 & 14.67 & 50.00 & 30.73 & 56.71 & 67.68 & 55.78 & 49.19 \\
\;\;+ TC + TG                & 2.56 & 50.00 & 24.43 & 9.83 & 49.80 & 28.00 & 55.15 & 59.76 & 59.08 & 50.07 \\
\;\;+ Rnd TC + TG            & \underline{35.00} & \underline{84.42} & \underline{57.78} & \underline{27.50} & \underline{71.00} & \underline{47.27} & \underline{59.62} & \underline{69.11} & 64.36 & \underline{50.66} \\

\rowcolor{gray!12}
\textbf{Rnd TC + Rnd TG} &
\textbf{47.78} & \textbf{95.84} & \textbf{69.94} &
\textbf{65.33} & \textbf{95.20} & \textbf{78.91} &
\textbf{60.98} & \textbf{70.93} & \underline{66.34} & \textbf{51.40} \\

\rowcolor{gray!6}
\textbf{$\Delta$ vs.\ base} &
{\color{ForestGreen}+42.56} & {\color{ForestGreen}+46.88} & {\color{ForestGreen}+44.55} &
{\color{ForestGreen}+53.00} & {\color{ForestGreen}+48.20} & {\color{ForestGreen}+50.82} &
{\color{ForestGreen}+1.90} & {\color{ForestGreen}+3.46} & {\color{red}-2.97} & {\color{ForestGreen}+2.94} \\

\bottomrule
\end{tabular}
}
\caption{\textbf{Sub-task Breakdown for Generalization Study.} 
Detailed results for VSPO (Navigation and Verification), VSP (Navigation and Verification), and WebMMU (Avg.\ = overall average, Act.\ = Agentic Action, Comp.\ = Visual Comprehension, Reason.\ = Multi-step Reasoning) from Table~\ref{tab:generalization_1_to_3}. 
\textbf{Rnd TC} and \textbf{Rnd TG} indicate randomized training on trajectories from 1-shot to 3-shot prompts. 
Best is \textbf{bold}, second-best is \underline{underlined}.}
\label{tab:generalization_full}
\end{table*}

\begin{table*}[t]
\centering
\small
\setlength{\tabcolsep}{5pt}
\renewcommand{\arraystretch}{1.05}

\begin{tabular}{
l
 S[table-format=2.2] S[table-format=2.2] S[table-format=2.2]
 S[table-format=2.2] S[table-format=2.2] S[table-format=2.2]
 S[table-format=2.2] S[table-format=2.2]
}
\toprule
\multirow{2}{*}{\textbf{Model}} &
\multicolumn{3}{c}{\textbf{VSPO}} &
\multicolumn{3}{c}{\textbf{VSP}} &
\multicolumn{2}{c}{\textbf{V*}} \\
\cmidrule(lr){2-4}\cmidrule(lr){5-7}\cmidrule(lr){8-9}
& {Nav} & {Verify} & {Overall}
& {Nav} & {Verify} & {Overall}
& {Attr.} & {Spatial.} \\
\midrule
\multicolumn{9}{c}{\textit{\textbf{Closed-Source Models}}} \\
\midrule
Gemini 2.5 flash & 15.44 & 68.96 & 40.12 & 34.50 & 76.40 & 53.55 & 74.78 & 59.21 \\
GPT 5 & 26.89 & 42.86 & 34.25 & 48.17 & 64.60 & 55.64 & 72.17 & 78.95 \\
Claude 4 sonnet & 37.56 & 67.92 & 51.56 & 48.17 & 66.00 & 56.27 & 59.13 & 60.53 \\
\midrule
\multicolumn{9}{c}{\textit{\textbf{Open-Source Models}}} \\
\midrule
Qwen 2.5 VL 3B & 5.67 & 50.91 & 26.53 & 7.50 & 49.80 & 26.73 & 38.26 & 52.63 \\
Qwen 2.5 VL 7B & 5.22 & 48.96 & 25.39 & 12.33 & 47.00 & 28.09 & 64.35 & 61.84 \\
Qwen 2.5 VL 32B & 7.56 & 53.12 & 28.56 & 24.33 & 45.40 & 33.91 & 72.17 & 72.37 \\
Qwen 2.5 VL 72B & 17.22 & 52.34 & 33.41 & 28.00 & 52.40 & 39.09 & \underline{79.13} & \textbf{81.58} \\
InternVL3 78B & 7.22 & 52.60 & 28.14 & 21.67 & 51.20 & 35.09 & \textbf{80.87} & \underline{81.58} \\
\midrule
\multicolumn{9}{c}{\textit{\textbf{Tool-Planning Models}}} \\
\midrule
Qwen 2.5 VL 7B + Tools & 11.89 & 48.96 & 28.98 & 17.83 & 45.60 & 30.45 & 49.57 & 63.16 \\
Qwen 2.5 VL 72B + Tools & 29.56 & 51.43 & 39.64 & 40.83 & 50.00 & 45.00 & 62.61 & 71.05 \\
GPT 5 + Tools & 38.11 & 69.87 & 52.75 & \underline{61.67} & \textbf{83.00} & \underline{71.36} & 66.96 & 75.00 \\
DeepEyes & 4.11 & 17.27 & 10.18 & 9.00 & 16.00 & 12.18 & 69.57 & 64.47 \\
PixelReasoner & 7.33 & 37.79 & 21.38 & 12.67 & 38.80 & 24.55 & 48.70 & 60.53 \\
\rowcolor{gray!12}
\textbf{AdaReasoner 7B} &
\textbf{47.78} & \underline{98.31} & \textbf{71.08} &
\textbf{63.00} & \underline{96.80} & \textbf{78.36} &
64.35 & 80.26 \\
\rowcolor{gray!6}
\textbf{$\Delta$ vs.\ base} &
{\color{ForestGreen}+42.56} & {\color{ForestGreen}+49.35} & {\color{ForestGreen}+45.69} &
{\color{ForestGreen}+50.67} & {\color{ForestGreen}+49.80} & {\color{ForestGreen}+50.27} &
0.00 & {\color{ForestGreen}+18.42} \\
\bottomrule
\end{tabular}

\caption{\textbf{Sub-task breakdown for Main Results.} 
Detailed results for VSPO (Navigation and Verification), VSP (Navigation and Verification), and V* (Attr.\ = Attribute Recognition, Spatial.\ = Spatial Relationship Reasoning) from Table~\ref{tab:part4_main_table}. 
Best is \textbf{bold}, second-best is \underline{underlined}.}
\label{tab:part4_full}
\end{table*}

\end{document}